\documentclass[sigconf]{acmart}

\usepackage{booktabs} 
\newcommand{\eat}[1]{}
\usepackage{bbold}
\usepackage{booktabs}
\usepackage{multirow}
\usepackage{balance}
\usepackage{siunitx}
\usepackage{bm}
\usepackage{algorithm}
\usepackage{transparent}
\usepackage{algpseudocode}

\usepackage{amsmath}
\usepackage{caption}
\usepackage{subcaption}
\usepackage{color}
\usepackage{url}
\usepackage{multirow}
\usepackage{enumerate}

\usepackage{amssymb}

\DeclareMathOperator*{\argmin}{min}
\algnewcommand{\Initialize}[1]{%
  \State \textbf{Initialize:}
  \Statex \hspace*{\algorithmicindent}\parbox[t]{.8\linewidth}{\raggedright #1}
}
\algnewcommand{\Inputs}[1]{%
  \State \textbf{Inputs:}
  \Statex \hspace*{\algorithmicindent}\parbox[t]{.8\linewidth}{\raggedright #1}
}
\algnewcommand{\Outputs}[1]{%
  \State \textbf{Outputs:}
  \Statex \hspace*{\algorithmicindent}\parbox[t]{.8\linewidth}{\raggedright #1}
}
\setcopyright{acmcopyright}

\copyrightyear{2019} 
\acmYear{2019} 
\acmConference[MM '19]{Proceedings of the 27th ACM International Conference on Multimedia}{October 21--25, 2019}{Nice, France}
\acmBooktitle{Proceedings of the 27th ACM International Conference on Multimedia (MM '19), October 21--25, 2019, Nice, France}
\acmPrice{15.00}
\acmDOI{10.1145/3343031.3350961}
\acmISBN{978-1-4503-6889-6/19/10}

\begin{document}
\title{Curiosity-driven Reinforcement Learning for Diverse Visual Paragraph Generation}
\eat{\titlenote{Produces the permission block, and
  copyright information}
\subtitle{Subtitle}
\subtitlenote{The full version of the author's guide is available as}
  \texttt{acmart.pdf} document}

\author{Yadan Luo$^\dag$, Zi Huang$^\dag$, Zheng Zhang$^\dag$, Ziwei Wang$^\dag$, Jingjing Li$^\ddag$, Yang Yang$^\ddag$} 
    \affiliation{ 
      $^\dag$The University of Queensland\\
      $^\ddag$University of Electronic Science and Technology of China
    }
    \email{lyadanluol@gmail.com, huang@itee.uq.edu.au, darrenzz219@gmail.com, ziwei.wang@uq.edu.au, lijin117@yeah.net, dlyyang@gmail.com}





\renewcommand{\shortauthors}{Luo, Y. et al}

\eat{Visual paragraph generation aims to automatically describe a given image from different perspectives and organize sentences in a coherent way. In spite of great advances achieved recently, existing paragraph captioning algorithms greedily focus on short-term phrase matching, thus highly degenerating the generalization of models and suppressing the diversity of expressions. In this paper, we propose a Curiosity-driven Reinforcement Learning (CRL) framework for diverse paragraph generation, where the visual-language policy network is optimized
Different from conventional REINFORCE algorithm that suffers from severe sparse and delayed rewards with long-sequence data like paragraphs, the derived curiosity module leverages the future state prediction error as dense intrinsic rewards to complement extrinsic rewards with respect to any caption evaluation metrics. Besides, without performing time-consuming warm-up for the policy networks in advance, discounted imitation learning is integrated into the proposed algorithm for efficient knowledge endowment. Extensive experiments conducted on the Standford image-paragraph dataset demonstrate the effectiveness and efficiency of the proposed method, improving performance by $38.4\%$ compared with state-of-the-arts. }

\begin{abstract}
Visual paragraph generation aims to automatically describe a given image from different perspectives and organize sentences in a coherent way. In this paper, we address three critical challenges for this task in a reinforcement learning setting: the mode collapse, the delayed feedback, and the time-consuming warm-up for policy networks. Generally, we propose a novel Curiosity-driven Reinforcement Learning (CRL) framework to jointly enhance the diversity and accuracy of the generated paragraphs. First, by modeling the paragraph captioning as a long-term decision-making process and measuring the prediction uncertainty of state transitions as intrinsic rewards, the model is incentivized to memorize precise but rarely spotted descriptions to context, rather than being biased towards frequent fragments and generic patterns. Second, since the extrinsic reward from evaluation is only available until the complete paragraph is generated, we estimate its expected value at each time step with temporal-difference learning, by considering the correlations between successive actions. Then the estimated extrinsic rewards are complemented by dense intrinsic rewards produced from the derived curiosity module, in order to encourage the policy to fully explore action space and find a global optimum. Third, discounted imitation learning is integrated for learning from human demonstrations, without separately performing the time-consuming warm-up in advance. Extensive experiments conducted on the Standford image-paragraph dataset demonstrate the effectiveness and efficiency of the proposed method, improving the performance by $38.4\%$ compared with state-of-the-art.

\end{abstract}


\maketitle
\fancyhead{}
\vspace{-0.5cm}

\section{Introduction}
With a rapid growth of multimedia data~\cite{yadan, yadan1}, understanding the visual content and interpreting it in natural language have been important yet challenging tasks, which could benefit a wide range of real-world applications, such as story telling \cite{visualstory,visualstory1,story2}, poetry creation \cite{poetry,poetry1,poetry2,poetry3} and support of the disabled. \begin{figure}[t]
\centering
        \includegraphics[width=0.48\textwidth]{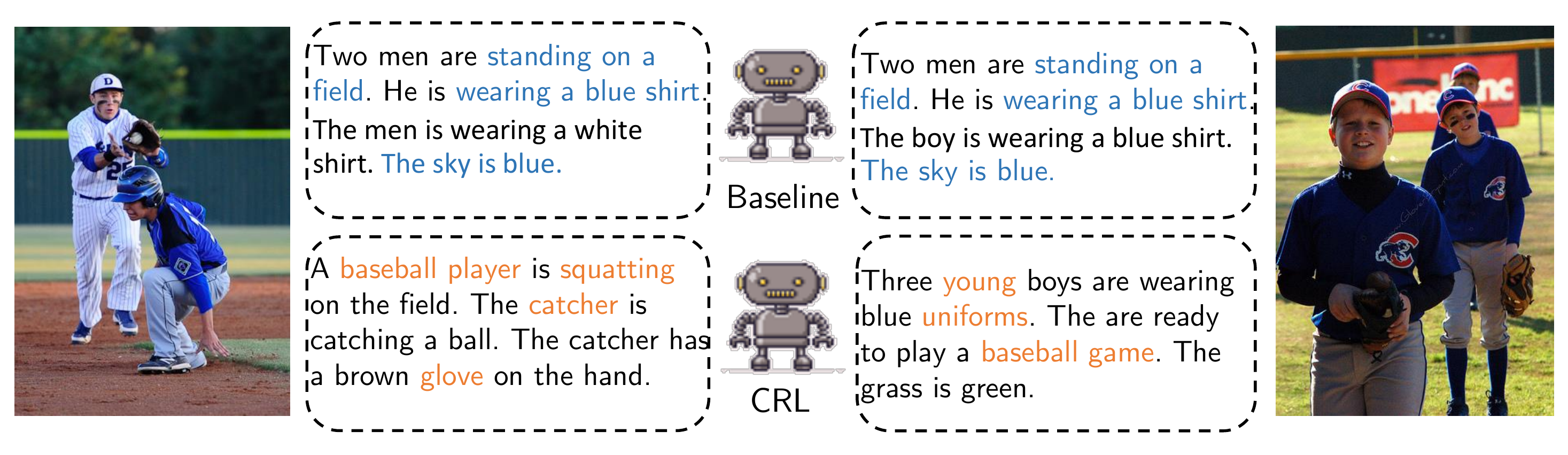}
\caption{An illustration of the mode collapse issue in paragraph captioning. Baseline model describes two distinct images with generic descriptions and similar patterns (as highlighted in blue), whereas the proposed CRL method generates diverse and more specific captions to context (as highlighted in orange).}
\label{fig:example}
\end{figure} While deep learning techniques have made remarkable progress in describing visual content via image captioning \cite{hanwang,SCST,ac,conGAN}, the obtained results are generally sentence-level, with fewer than twenty words. Generating such a short description could hardly convey adequate messages for all subtle objects and relationships, to say nothing of further pursuing coherence and diversity. Consequently, it is more natural to depict images in paragraph format \cite{video, para, im2p, RTTGAN, eccv}, which has been investigated recently. 

Generally, existing mainstream paragraph captioning models \cite{im2p} follow the encoder-decoder architecture, where the language decoder is fully supervised to maximize the posterior probability of predicting each word, given the previous ground-truth sequence and image representations extracted by the visual encoder. The word-level cross-entropy objective, in this way, will encourage the use of exactly the same \textit{n}-grams that appear in ground-truth samples, making the paragraph captions lack completeness and consistency. Motivated to achieve more diverse and natural descriptions, an emerging line of work \cite{RTTGAN, conGAN, eccv} combines supervised learning with generative adversarial models~\cite{RTTGAN,conGAN} or auto-encoders~\cite{eccv}, aiming to capture inherent ambiguity of captions with a low-dimensional Gaussian manifold and model the structure of paragraphs in a hierarchical way. Nevertheless, existing paragraph captioning approaches are far from optimal due to two major issues, \textit{i.e.,} the \textit{mode collapse} and the \textit{exposure bias}. First, the simple Gaussianity assumption is not sufficient enough to fully preserve the ground-truth distribution, many modes of which are underrepresented or missing. For instance, given two distinct pictures of baseball games as Figure \ref{fig:example} shows, the generated paragraphs by the baseline model only describe objects and actions with vague and general words (e.g., ``two men'', ``standing'') rather than specific noun entities (e.g., ``baseball player'', ``catcher'') or vivid verbs (e.g., ``squatting''). Second, the language decoder predicts based on different inputs during training and testing, i.e., the ground-truth sub-sequences during training yet its own predictions during testing. This discrepancy as exposure bias severely hurts the model performance.

Recently, another line of work tackles the exposure bias and takes advantage of non-differential evaluation feedback by applying reinforcement learning, especially the REINFORCE \cite{reinforce} algorithm for the sentence-level captioning task \cite{hanwang,ac,SCST,SPIDEr,temporal}. This strategy reformulates the image captioning as the sequential decision-making process, where the language policy based on its previous decisions is directly optimized. Besides, it is more reasonable to achieve long-term vision and less greedy behaviors by optimizing the sentence-level evaluation metrics such as BLEU~\cite{bleu}, METEOR~\cite{meteor} and CIDEr \cite{cider} instead of the cross entropy loss. 

Unfortunately, it is challenging to extend the success to the paragraph captioning task: (1)\textbf{ \textit{Mode Collapse:}}
optimizing evaluation metrics still does not alleviate the fixed-pattern issue, where the strategy could be easily tricked by repetition of frequent phrases and generic representations, yielding less variety of expressions; (2)\textbf{ \textit{Delayed Feedback:}} the language policy only receives feedback from the evaluator when the entire sequence is produced, resulting in high training variance especially with long sequence data like paragraphs; (3)\textbf{ \textit{Time-consuming Warm-up:}} current reinforcement learning suffers from low sample efficiency, which causes an unbearable time and computational cost for trial and error. Therefore, it usually requires long-term supervised pre-training for policy warm-up.

To address above-mentioned issues, in this paper, we propose a novel Curiosity-driven Reinforcement Learning (CRL) framework for diverse visual paragraph generation. Firstly, we design an intrinsic reward function (the curiosity module) that encourages the policy to explore uncertain behaviors and visit unfamiliar states, thereby increasing the expression variety rather than pursuing local phrase matching. In particular, the curiosity module consists of two sub-networks on top of language policy networks with self-supervision, \textit{i.e.,} the Action Prediction Network (AP-Net) and the State Prediction Network (SP-Net). Moreover, the SP-Net measures state prediction error at each time steps as dense intrinsic rewards, which are complementary to the delayed extrinsic reward. Different from conventional reinforcement learning~\cite{SCST,hanwang} that simply averages the extrinsic reward to each state, we further adopt the temporal difference learning~\cite{sutton} to correct the extrinsic reward estimation, considering the correlations of successive actions. Lastly, to avoid time-consuming warm-up for policy networks, our algorithm seamlessly integrates discounted imitation learning to stabilize learning for fast convergence, and gradually weakens supervision signal without shrinking action space.

Overall, our contributions can be briefly summarized as follows:
\begin{itemize}
    \item To our best knowledge, this is the first attempt to tackle the visual paragraph generation problem with pure reinforcement learning. Different from the conventional REINFORCE algorithm, our CRL learning motivates visual reasoning and language decoding both intrinsically and extrinsically.
    \item The intrinsic curiosity complements sparse and delayed extrinsic rewards with prediction error measurement, guiding the agent to fully explore and achieve a better policy.
    \item Instead of pre-training policy networks with supervised learning, we jointly stabilize reinforcement learning with discounted imitation learning for fast convergence.
    \item We show the effectiveness of the proposed strategy through extensive experiments on the Standford paragraph captioning benchmark and demonstrate the diversity of generated paragraphs with a visualization of semantic network graphs. 
\end{itemize}
\eat{The rest of paper is organized as follows. Section 2 presents a brief review of reinforced image captioning methods and intrinsically motivated RL. Section 3 introduces the details of the proposed paragraph captioning model. The experimental comparisons with state-of-the-art and ablation study are presented in Section 4, followed by the conclusions in Section 5.}

\section{Related Work}
\subsection{Sentence-level Captioning with Reinforcement Learning}
 Inspired by the recent advances in reinforcement learning, several attempts have been made to apply policy gradient algorithms to image captioning task~\cite{yang, yang1, yang2}, which could generally be categorized into two groups: policy based and actor-critic based. Policy based methods (e.g., DISC \cite{conGAN}, SCST \cite{SCST}, PG-SPIDEr \cite{SPIDEr}, CAVP \cite{hanwang}, TD \cite{temporal}) utilize the unbiased REINFORCE \cite{reinforce} algorithm which optimizes the gradient of the expected reward by sampling a complete sequence from the model during training. To suppress high variance of Monte-Carlo sampling, Self-critical Sequential Training (SCST) \cite{SCST} utilizes a baseline subtracted from the return which is added to reduce the variance of gradient estimation. Rather than obtaining a single reward at the end of sampling, actor-critic based algorithms (e.g., Embedded Reward \cite{ER}, Actor-Critic \cite{ac}, Adapt \cite{adapt}, HAL \cite{HRL}) learn both a policy and a state-value function (``crtic''), which is used for bootstrapping, \textit{i.e.,} updating a state from subsequent estimation, to reduce variance and accelerate learning \cite{sutton}. Different from existing work, the proposed CRL algorithm learns about a critic from the inner environment, complementing the extrinsic reward from the perspective of agent learning.
\vspace{-0.2cm}
\subsection{Paragraph-level Captioning}
While sentence-level captioning has been extensively studied, the problem of generating paragraph-level descriptions still remains under-explored. Existing solutions include (1) generating sentences individually with detected region proposals (DenseCap \cite{densecap}), or with topic learning via the Latent Dirichlet Allocation (TOMS \cite{toms}); (2) preserving semantic content and linguistic order with hierarchical structure (Region-Hierarchical \cite{im2p}). To further encourage the coherence and naturalness among successive sentences, this model was further extended by Liang et al. \cite{RTTGAN} and Dai et al. \cite{conGAN} by adopting adversarial learning. To tackle the training difficulties of Generative Adversarial Networks (GANs), Chatterjee et al. \cite{eccv} modeled the inherent ambiguity of paragraphs via a variational auto-encoder formulation. From another perspective, Wang et al. \cite{DAM} leveraged depth estimation to discriminate objects at various depths and capture subtle interactions.

 \subsection{Intrinsically Motivated Reinforcement Learning}
In reinforcement learning area, much theoretical work has been done on improving agent exploration and shaping sparse rewards via intrinsic motivation \cite{intrinsic}, where an information-theoretic critic measures the agent's surprisal \cite{surprisal, curiosity, curiosity1, RND} (based on prediction error) or state novelty \cite{novelty, novelty1, novelty2, novelty3} (based on counts of visited states), motivating the agent from the inner environment. Our proposed algorithm shares the same spirit with the former group. But instead of testing on simulated games, we, \textit{for the first time}, adapt the intrinsic reward and validate its effectiveness and efficiency on a more practical task, \textit{i.e.,} paragraph captioning.

\begin{figure}[t]
\centering
        \includegraphics[width=0.48\textwidth]{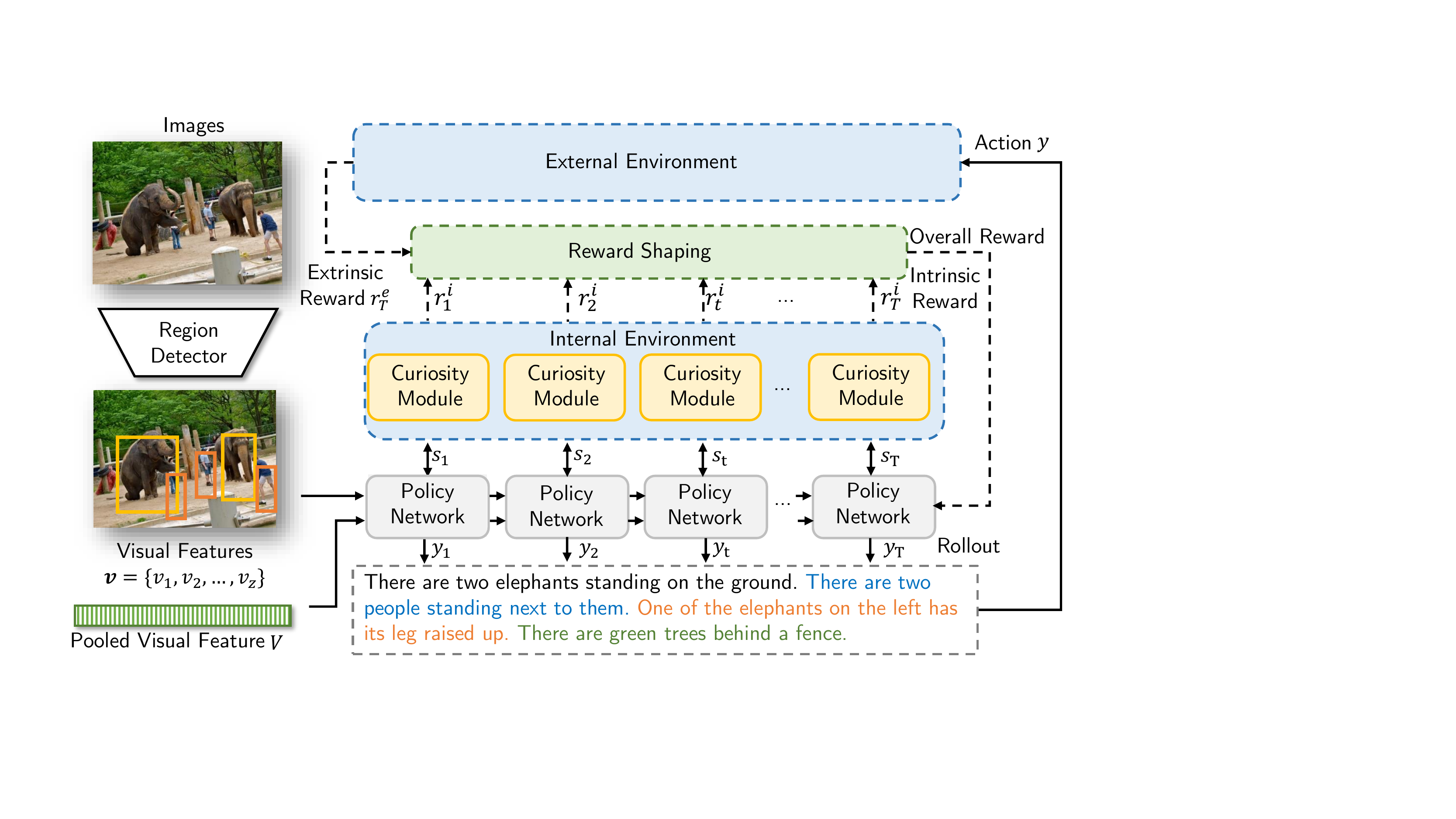}
\caption{The general flowchart of the proposed image paragraph captioning model.}
\label{fig:flowchart}
\end{figure}


\section{Curiosity-driven Learning}
The overview of the proposed paragraph captioning framework is illustrated in Figure \ref{fig:flowchart}. We firstly formulate the task of visual paragraph generation, followed by the introduction of the language policy network and policy learning. To enhance the agent exploration, two sub-networks of the curiosity module are trained in a self-supervised manner: the state prediction network (SP-Net) and the action prediction network (AP-Net). Then, we explain the detailed calculation for rewards with temporal-different learning, discounted imitation learning and collaborative optimization for all objectives.

\subsection{Problem Formulation} \label{PF}
Our target is to generate the paragraph caption $ y = \{ y_1,  y_2, \ldots,  y_T\}\in \{1,0\}^{T\times D}$ for any given image $I$, where $T$ denotes the length of the generated caption and $D$ is the vocabulary size. The proposed framework follows the general encoder-decoder structure (see Figure \ref{fig:flowchart}), where  $E$-dimensional visual features for $m$ local regions $ v = \{ v_1,  v_2, \ldots,  v_{m}\}\in\mathbb{R}^{m\times E}$ are extracted by the Faster RCNN~\cite{fastercnn} at the encoding stage. Language sequences are decoded by Recurrent Neural Networks (RNN) step by step. Different from traditional settings that directly force the language decoder to mimic ground truths by using cross entropy loss, our work casts the problem in reinforcement learning in order to optimize non-differential evaluation metrics (\textit{e.g.,} CIDEr) and suppress the exposure bias. Playing as the ``policy'' $\pi_{\theta}$ in a finite Markov decision process (MDP), the language decoder model parameterized by $\theta$ predicts the next word $y_t$ (``action'' $a_t$) at time $t$ based on the hidden ``state'' $s_t$. To suppress the sparsity and delay issues of the extrinsic reward $r^e$, the derived curiosity critic predicts the expected ``intrinsic reward'' $r^i_t$ at each time step, which greatly complements and shapes the final reward.

\subsection{Visual-Language Policy Network}\label{PN}
To adaptively control visual signals and generate context-aware descriptions, we adopt a double-layer LSTM structure coupled with attention mechanism as the policy networks (see Figure \ref{fig:flowchart}). The first LSTM layer serves as a top-down visual attention model, taking the input word $y_{t-1}$, concatenated with the mean-pooled image features $V$ and previous $Z$-dimensional state of the language LSTM $s_{t-1}^{lang}$ at each time step. Therefore, the state transition for the visual LSTM is,
\begin{equation}
\begin{split}
     s_t^{vis} = \text{LSTM}([ s^{lang}_{t-1},  W_v^{T} V,  W_e^{T} y_{t-1}],  s_{t-1}^{vis}),
\end{split}
\end{equation}
where $W_e$, $W_v$ are the learnable weights and $w_t\in\{0, 1\}^{D}$ is one-hot embedding of the input word at the time step $t$. To further attend local visual features $v$ based on language policy, the weighted visual features $\hat{v}_t\in\mathbb{R}^{E}$ can be calculated as,
\begin{equation}
    \begin{split}
     \hat{v}_t =  \sum_{i=1}^m \text{softmax}( W_{\alpha}^T\text{tanh}( W_v  v_i +  W_h s_t^{vis})) v_{i},
    \end{split}
\end{equation}
where $W_v$, $W_h$, and $W_\alpha$ are the weights to be learned. Obtaining weighted visual features and the hidden state $s^{vis}_t$ from the attention LSTM, the top-level language LSTM gives the conditional distribution $\pi_{\theta}(y_t|\cdot)$ for the next word $y_t$ prediction, \textit{i.e.,}
\begin{equation}
    \begin{split}
        \pi_{\theta}(y_t|s_t) &= \text{softmax}( W_p s_t^{lang}),\\
        s_t^{lang} &= \text{LSTM}([\hat{v}_t, s_t^{vis}], s_{t-1}^{lang}),
    \end{split}
\end{equation}
where $W_p$ is the learnable weight. To encourage the agent to explore rare attended areas and words, we concatenate the state $s_t = [s_t^{vis}, s_t^{lang}]$, which will be used in policy learning and discounted imitation learning. To form the complete paragraph, the distribution can be calculated as the dot product of conditional distributions from previous steps,
\begin{equation}
    \pi_{\theta}(y_{1:T}) = \prod_{t=1}^T\pi_{\theta}(y_t|s_t).
\end{equation}

\subsection{Policy Learning}\label{RL}

In reinforcement learning, the policy network $\pi_\theta$ leverages the experiences obtained from interacting with an environment to learn behaviors that maximize a reward signal. Generally, the RL loss can be presented as,
\begin{equation}
\argmin_{\theta}\mathcal{L}_{RL} = -\mathbb{E}_{y\sim \pi_{\theta}}[ A^{\pi_{\theta}}(s, y)],
\end{equation}
where $A^{\pi_{\theta}}(s, y) = Q^{\pi_{\theta}}(s, y) + V^{\pi_{\theta}}(s)$ is the advantage function. $Q^{\pi_{\theta}}(s, y)$ stands for the state-action function estimating the long-term value instead of the instantaneous reward, and $V^{\pi_{\theta}}(s)$ indicates state value function, which serves as the inner critic. The core idea is to incentivize the policy to increase the probability of actions that are correct and rarely-seen. $s=\{s_1,s_2,\ldots,s_T\}$ denotes the concatenated hidden states of the policy network. Based on policy gradient \cite{sutton}, the gradient of non-differentiable reward-based loss function can be derived as,

\begin{equation}
    \begin{split}
        \nabla_\theta\mathcal{L}_{RL} &= -\mathbb{E}_{y\sim \pi_{\theta}}[ A^{\pi_{\theta}}(s, y)\nabla_\theta\log \pi_\theta(y|s)]\\
        &= - \sum_{t=1}^{T}A^{\pi_{\theta}}(s_t, y_t)\nabla_\theta\log \pi_\theta(y_t|s_{t}).
    \end{split}
\end{equation}
\
\subsection{Self-supervised State Prediction (SP-Net)}\label{SP}
Before estimating the advantage function, we first detail two sub-networks for the state value function $V^{\pi_{\theta}}(s)$. The SP-Net is trained to predict the future state embedding $\phi(s_{t+1})$ based on the input action $y_t$, and $\phi(s_t)$, where $\phi(\cdot)$ indicates the state embedding layer and helps filter irrelevant memory for the prediction of next state. The mean-squared error is used as the objective function for SP-Net,
\begin{equation}
\begin{split}
\argmin_{\theta, \theta_{SP}, \theta_{\phi}}\mathcal{L}_{SP} &= \frac{1}{2}\|\tilde{\phi}(s_{t+1}) - \phi(s_{t+1})\|_2^2\\
\tilde{\phi}(s_{t+1}) &= \mathcal{G}(\phi(s_t), y_t;\theta_{SP}),
\end{split}
\end{equation}
where $\mathcal{G}(\cdot; \theta_{SP})$ denotes the nonlinear transformation of SP-Net parameterized by $\theta_{SP}$. In this way, the state value function can be obtained as,
\begin{equation}\label{v}
    V^{\pi_{\theta}}(s) = \frac{\rho}{2}\sum_{t=1}^T\|\tilde{\phi}(s_{t}) - \phi(s_{t})\|^2_2,
\end{equation}
where $\rho$ is the hyper-parameter. The prediction error quantifies the agent uncertainty towards the environment. The policy network trained to maximize state prediction error will explore transitions with less experience and high confusion, therefore rare attended areas and infrequent expressions can be well captured. 

\subsection{Self-supervised Action Prediction (AP-Net)}\label{AP}
Given the transition tuple $(s_t, s_{t+1}, y_t)$, the action prediction network targets at predicting the action $y_t$ based on state transition.
The objective of AP-Net can be defined as,
\begin{equation}
\begin{split}
\argmin_{\theta, \theta_{AP}, \theta_{\phi}}\mathcal{L}_{AP} &= -\sum_{t=1}^Tq(y_t)\log(\tilde{y_t})\\
\tilde{y_t} &= \mathcal{F}(\phi(s_t), \phi(s_{t+1});\theta_{AP}),
\end{split}
\end{equation}
where $\tilde{y_t}$ is the prediction of current action, shown as a softmax distribution among all possible words. $q(y_t)$ is the real distribution of action $y_t$ and $\mathcal{F}(\cdot)$ denotes the nonlinear transformation of AP-Net parameterized by $\theta_{AP}$. The intuition of AP-Net is to learn state embedding that corresponds to meaningful patterns of human writing behaviors, suppressing the impact of outliers.

\begin{figure}[t]
\centering
        \includegraphics[width=0.3\textwidth]{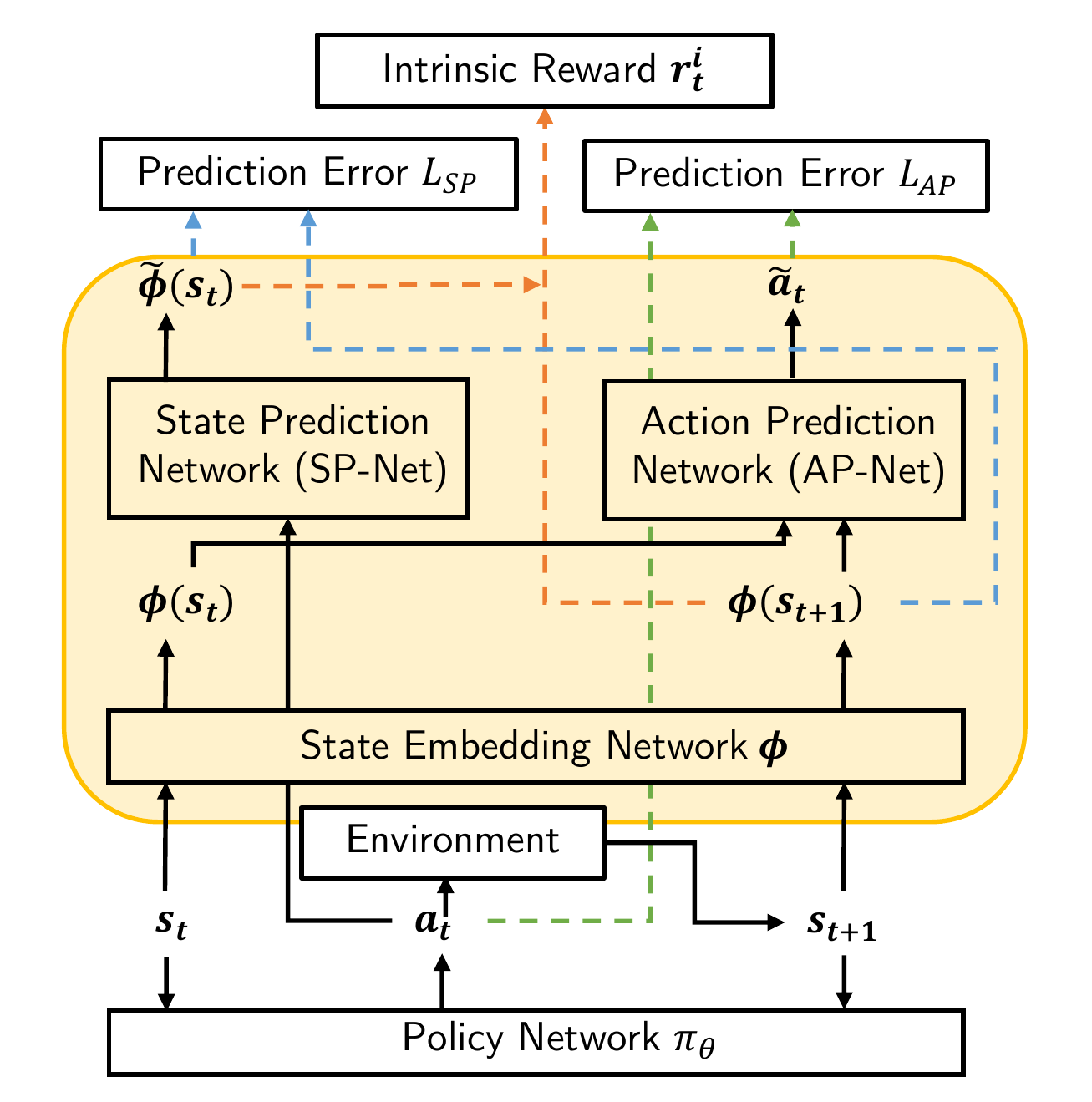}
        \vspace{-0.2cm}
\caption{An illustration of the curiosity module. The blue and the green dotted line denote the loss calculation for $\mathcal{L}_{SP}$ and $\mathcal{L}_{AP}$, respectively. The orange line shows the calculation for the intrinsic reward $r_t^i$.}
\label{fig:ICM}
\end{figure} 

\subsection{Reward Shaping}\label{reward}
To encourage an agent to explore its environment for acquiring new knowledge and guide it to generate accurate and diverse paragraphs, the overall reward is generated by two parts, dense intrinsic curiosity reward $r^i$ and sparse extrinsic reward $r^e$. The policy network is expected to maximize the weighted sum of two rewards.
\subsubsection{Extrinsic Reward}
To improve the fidelity and interpretability of the learned paragraph, the extrinsic reward $r^e_t$ is refined as the linear combination of linguistic measures. Specifically, we select the most representative and commonly used metrics, \textit{i.e., } BLEU-4 \cite{bleu} and CIDEr \cite{cider},
\begin{equation}
    r_{t}^e = \begin{cases}
    a\cdot\text{BLEU-4}(y_{1:T}) + b\cdot\text{CIDEr}(y_{1:T}), & \text{if $t=T$}\\
    0, & \text{otherwise}.
    \end{cases}
\end{equation}
In our case, the hyper-parameters $a$ and $b$ are empirically set to $1$ and $2$, respectively. Under such a reward setting, we adopt the temporal-difference learning TD($\lambda$) \cite{sutton} to estimate the action-state function $Q^{\pi_{\theta}}(s_t, y_t)$ for each time step, 
\begin{equation}
\begin{split}
    G_{t:t+j} = r^e_t + \gamma r^e_{t+1}+\gamma^2 r^e_{t+2} +\ldots +\gamma^{j}r^e_{t+j}\\
    Q^{\pi_{\theta}}(s_t, y_t) = (1-\lambda)\sum_{j=0}^{T-t}\gamma^{j}G_{t:t+j} + \lambda^{T-t}G_t,
\end{split}
\end{equation}
where the $j$-step expected return $G_{t:t+j}$ is defined as the sum of expected future rewards from the next $j$ steps. The $\lambda$ indicates the trade-off parameter between the future estimation and the current estimation. The discounted factor $\gamma$ enables variance reduce by down-weighting extrinsic rewards. For simplicity, we set $\lambda$ to 1, and the overall $Q$ function can be formulated as,
\begin{equation}\label{q}
    Q^{\pi_{\theta}}(s, y) = \sum_{t=1}^T\sum_{j=0}^{T-t}\gamma^{j}r^e_{t+j} = \sum_{t=1}^T\gamma^{T-t}r^e_T,
\end{equation}
\subsubsection{Intrinsic Reward}
As discussed in Section \ref{SP}, we train a state prediction network and calculate the prediction error as intrinsic reward $r^i = V^{\pi_{\theta}}(s)$ (see Equation \eqref{v}). Therefore, the gradient of policy network can be rewritten as,
\begin{align}
        \nabla_\theta\mathcal{L}_{RL} =& - \sum_{t=1}^{T}(Q^{\pi_{\theta}}(s, y)+V^{\pi_{\theta}}(s))\nabla_\theta\log \pi_\theta(y_t|s_{t})\\
        =& - \sum_{t=1}^{T}(\gamma^{T-t}r^e_T+\frac{\rho}{2}\|\tilde{\phi}(s_{t}) -\phi(s_{t})\|^2_2)\nabla_\theta\log \pi_\theta(y_t|s_{t}).\nonumber
\end{align}

\subsection{Discounted Imitation Learning}\label{imitation}
A major challenge of the reinforced agent to have a good convergence property is that the agent must start with a good policy at the beginning stage. The low sample efficiency issue \cite{efficiency} causes a huge amount of time and computational cost for trial and error. Existing sentence-level captioning methods \cite{SCST,hanwang,SPIDEr,ac} with reinforcement learning apply the cross entropy loss to the language decoder for warm up, which is defined as,
\begin{equation}
\begin{split}
   \argmin_{\theta}\mathcal{L}_{XE} &= -\sum_{t=1}^T\log \pi_{\theta}(y_t^*|y_{t-1}^*,\ldots,y_1^*,V),
\end{split}
\end{equation}
where $y^* = \{y_1^*, y_2^*, \ldots, y_T^*\}$ is the human-labeled ground-truth. Despite that supervised learning is essential to initialize the policy network, it usually consumes a long period of time (e.g., 40 epochs on Standford dataset), and highly restricts the agent search space, which probably leads to a  local minimum. Therefore, we introduce the discounted imitation learning at the first epoch of training, then gradually decrease the loss coefficient $\eta$ to weaken the supervision.

\subsection{Collaborative Optimization}\label{end}
To collaboratively optimize the objectives of reinforcement learning, the curiosity modules and discounted imitation learning, the overall learning loss function can be formulated as,
\begin{equation}
\begin{split}
     \argmin_{\theta,\theta_{AP},\theta_{SP},\theta_{\phi}} \mathcal{L}_{RL} + \alpha\mathcal{L}_{AP} + \beta\mathcal{L}_{SP} + \eta\mathcal{L}_{XE},
\end{split}
\end{equation}
where $\alpha$ and $\beta$ are constant loss coefficients, $\eta$ is a dynamic scaling factor that gradually reduces with ten percent decay every epoch. Notably, we dynamically estimate the intrinsic reward of agent behavior to shape reward signal, which avoids additional baseline calculation in the advantage function. The overall algorithm is shown in the Algorithm \ref{alg:1}.
\begin{algorithm}[!h]
	\begin{algorithmic}[1]
		\Inputs{Annotated image-paragraph set $\mathcal{D}=\{I_{(n)}, y^{*}_{(n)}\}_{n=1}^N$;}
		\Outputs{Visual-language policy network $\pi_{\theta}$;}
		\Initialize{Hyper-parameters: $\rho$, $\gamma$, $\alpha$, $\beta$, $\eta$, $a$, $b$;\\
		Visual features $v$ from $m$ regions;\\
		Minibatch size $m$ and learning rate $\mu$;}
        \For{k epochs}
        \For{T time steps}
        \State Sample action $a_t$ based on policy $\pi_{\theta}$
        \EndFor
        \State Calculate intrinsic rewards $\{r^i_t\}_{t=1}^{T}$ and state-value function $V^{\pi_{\theta}}(s)$ in Equation \eqref{v};
        \State Calculate extrinsic rewards $\{r^e_t\}_{t=1}^{T}$ and action-state function $Q^{\pi_{\theta}}(s, y)$ in Equation \eqref{q};
        \State Update parameters $\theta$, $\theta_{SP}$, $\theta_{AP}$ and $\theta_{\phi}$ by descending stochastic gradients:
        \State $\theta\gets\theta-\mu\cdot\nabla_\theta\frac{1}{m}(\mathcal{L}_{RL}+\alpha\mathcal{L}_{AP}+\beta\mathcal{L}_{SP}+\eta\mathcal{L}_{XE})$;
         \State $\theta_{AP}\gets\theta_{AP}-\mu\cdot\nabla_{\theta_{AP}}\frac{1}{m}\mathcal{L}_{AP}$;
          \State $\theta_{SP}\gets\theta_{SP}-\mu\cdot\nabla_{\theta_{SP}}\frac{1}{m}\mathcal{L}_{RL}$;
          \State $\theta_{\phi}\gets\theta_{\phi}-\mu\cdot\nabla_{\theta_{\phi}}\frac{1}{m}(\alpha\mathcal{L}_{AP}+\beta\mathcal{L}_{SP})$;
        \State update the dynamic factor $\eta\gets\delta\eta$;
        \EndFor
	\end{algorithmic}
	\caption{Pseudo-code of the Proposed CRL Learning.}
	\label{alg:1}
\end{algorithm}

\section{Experiments}
\subsection{Settings}
\subsubsection{Dataset}
All state-of-the-art methods and our proposed method are evaluated on the \textit{Stanford image-paragraph }dataset \cite{im2p}, where 14,579 image-paragraph pairs from the Visual Genome and MS COCO dataset are used for training, 2,490 for validation and 2,492 for testing. The number of unique words in its vocabulary is 12,186. All images are annotated with human-labeled paragraphs of 67.5 words on average.  
\subsubsection{Evaluation Metrics}
We report the performance of all models on six widely used automatic evaluation metrics, i.e., \textbf{BLEU-\{1,2,3,4\}} \cite{bleu}, \textbf{METEOR} \cite{meteor} and \textbf{CIDEr} \cite{cider}. BLEU-$n$ is defined as the geometric mean of $n$-gram precision scores and CIDEr measures $n$-gram accuracy by term-frequency inverse-document-frequency (TF-IDF). METEOR is defined as the harmonic mean of precision and recall of exact, stem, synonym, and paraphrase matches between paragraphs.

\begin{table*}[t]
	\caption{Performance comparisons using BLEU-\{1,2,3,4\}, METEOR and CIDEr on Standford Image-paragraph dataset. The human performance is provided for reference. In our setting, the proposed CRL only optimizes CIDEr and BLEU-4 and achieves the highest scores compared with state-of-the-art.}
	\label{tab:comparison}
    \centering
    \resizebox{0.8\textwidth}{!}{%
	\begin{tabular}{l|c|c|*{6}{c}}
	    \toprule
		\toprule
		\textbf{Methods} &\textbf{Language Decoder}	&\textbf{Beam Search}	& \textbf{METEOR} & \textbf{CIDEr} & \textbf{BLEU-1} & \textbf{BLEU-2} & \textbf{BLEU-3} & \textbf{BLEU-4}\\
		\hline		
		Sentence-Concat (Neuraltalk\cite{Karpathy_2015_CVPR}) &1*LSTM &2-beam &12.05 & 6.82 & 31.11 & 15.10 & 7.56 &3.98\\
		Sentence-Concat (NIC\cite{NIC}) &1*LSTM &2-beam &9.27 & 7.09 & 22.31 & 10.72 & 4.91 & 2.32\\
		\hline
		Image-Flat (NIC\cite{NIC}) &1*LSTM &2-beam & 13.44 & 14.71 & 34.80 & 19.42 & 10.91 & 6.03\\
		DAM-Att (\cite{DAM}) &2*LSTM &Greedy & 13.91 & 17.32 &35.02 & 20.24 &11.68 & 6.57\\
		TOMS (\cite{toms})&1*LSTM &3-beam &18.60 &20.80 &43.10 & 25.80 &14.30 &8.40 \\
		\hline
		Region-Hierarchical (\cite{im2p}) &2*LSTM &2-beam & 13.85 & 10.64 & 35.58 & 17.94 &9.08 & 4.49\\
        RTT-GAN (\cite{RTTGAN})  &2*LSTM &2-beam & 17.12 &16.87 &41.99 &24.86 &14.89 &9.03\\
        VAE (\cite{eccv})  &2*GRU &Greedy &\textbf{18.62} &20.93 &42.38 &25.52 &15.15 &9.43\\
        SCST (\cite{SCST})  &1*LSTM &Greedy &16.01 &22.74 &40.89 &23.71 &14.43 &8.38\\
        \hline
        \textbf{CRL}  &1*LSTM &Greedy &17.71 &\textbf{25.03} &\textbf{43.10} &\textbf{26.93} &\textbf{16.65} &\textbf{9.91}\\
        \textbf{CRL}  &1*LSTM &2-beam &17.42 &\textbf{31.47} &\textbf{43.12} &\textbf{27.03} &\textbf{16.72} &\textbf{9.95}\\
		\hline
		Humans (as in \cite{im2p})  &- &- & 19.22 & 28.55 & 42.88 & 25.68 & 15.55 & 9.66\\
		\bottomrule
		\bottomrule
	\end{tabular}
    }
\end{table*}

\subsection{Baselines}
We compare our approach with several state-of-the-art paragraph captioning methods and one RL-based method. 

\textit{\textbf{Sentence-Concat:}} Two sentence-level captioning models (\textbf{Neu}- \textbf{raltalk} \cite{Karpathy_2015_CVPR} and \textbf{NIC} \cite{NIC}) pre-trained on the MS COCO dataset are adopted to predict five sentences for each given image, which are further concatenated into a paragraph. 

\textit{\textbf{Image-Flat:}} Different from sentence-concat group, \textbf{Image-Flat} \cite{NIC} method directly generates a paragraph word by word, with the ResNet-152 network \cite{resnet} for visual encoding and a single LSTM layer to decode language. \textbf{DAM-Att} \cite{DAM} couples the encoder-decoder architecture with attention mechanism, and additionally introduces depth information to enhance recognition to spatial object-object relationships. \textbf{TOMS} \cite{toms} learns topic-transition among multiple sentences with Latent Dirichlet Allocation (LDA).

\textit{\textbf{Hierarchical:}} 
\textbf{Region-Hierarchical} \cite{im2p} leverages a hierarchical recurrent network to learn sentence topic transition and decode language sentence by sentence. \textbf{RTT-GAN} \cite{RTTGAN} implements the hierarchical learning in a GAN \cite{GAN} setting, where the generator mimics the human-annotated paragraphs and tries to fool the discriminator. \textbf{VAE} \cite{eccv} models the paragraph distribution with variational auto-encoder \cite{VAE}, which preserves the coherence and global topics of paragraphs. Notably, Liang et al. took advantage of the local phrases that are predicted by the dense-captioning model  \cite{densecap}, which additionally used training data from the MS-COCO dataset.

\textit{\textbf{REINFORCE:}} For fair comparison, we compare the proposed framework with the RL-based image captioning method \textbf{SCST} \cite{SCST}. The model shares the same backbone encoder-decoder structure but a different reinforcement learning strategy and reward functions. As it requires supervised warm-up for policy networks, we pre-train the SCST with the cross-entropy (XE) objective using ADAM optimizer \cite{adam} with the learning rate of $5\times10^{-4}$.

\begin{table*}[t]
\begin{center}
\caption{Ablative performance comparisons on Standford image-paragraph dataset. ``w/o'' indicates without. The best performances are shown in boldface.} \label{ablation}
\resizebox{0.9\textwidth}{!}{
\begin{tabular}{|c|c|cccccc|cccccc|}
\hline
\multirow{2}{*}{\textbf{Methods}}&
\multirow{2}{*}{\textbf{Policy}}&
\multicolumn{6}{c|}{\textbf{ResNet Features}}&
\multicolumn{6}{c|}{\textbf{Region Features}}\\
\cline{3-14}
& &METEOR &CIDEr &BLEU-1 &BLEU-2 &BLEU-3 &BLEU-4
&METEOR &CIDEr &BLEU-1 &BLEU-2 &BLEU-3 &BLEU-4\\
\hline
\multirow{3}{*}{CRL w/o RL} &FC &13.10 &11.18 &36.19 &17.67 &8.41 &3.93
&13.33 &11.83 &37.01 &18.32 &8.77 &4.35
\\
&Att &13.77 &12.34 &37.40 &20.94 &11.48 &6.10
 &13.35 &12.23 &36.50 &19.11 &9.28 &4.39
\\
 &Up-Down &14.02 &11.46 &37.68 &19.17 &9.34 &4.34
&14.28 &14.10 &38.07 &20.42 &10.57 &5.26

\\
\hline
\multirow{3}{*}{CRL w/o intrinsic} &FC &15.12 &18.31 &39.38 &21.81 &11.84 &6.23 &15.67 &20.14 &40.98 &23.04 &13.44 &7.58\\
&Att &15.89 &19.07 &41.32 &24.72 &14.04 &7.96 &15.92 &19.12 &40.31 &24.82 &14.36 &8.27\\
 &Up-Down &15.91 &20.45 &41.41 &24.77 &14.40 &8.14 &16.01 &22.74 &40.89 &23.71 &14.43 &8.38\\
\hline
\multirow{3}{*}{CRL}&FC &15.53 &20.31 &39.68 &22.66 &12.69 &6.89 &15.87 &21.13 &40.98 &24.30 &14.12 &7.96
\\
 &Att &16.13 &19.67 &41.17 &24.18 &14.92 &8.92
&16.11 &19.21 &41.17 &25.00 &15.03 &8.82
\\
 &Up-Down &\textbf{16.71} &\textbf{24.99} &\textbf{41.88} &\textbf{25.24} &\textbf{15.25} &\textbf{9.03}
&\textbf{17.71} &\textbf{25.03} &\textbf{43.10} &\textbf{26.93} &\textbf{16.65} &\textbf{9.91}
\\
\hline
\end{tabular}}
\end{center}
\end{table*}

\subsection{Implementation Details} 
Our source code is based on PyTorch \cite{pytorch} and all experiments are conducted on a server with two GeForce GTX 1080 Ti GPUs.
\subsubsection{Data Pre-processing}~\label{dp}
For textual pre-processing, we first tokenize all annotated paragraphs, and replace words that appear less than five times with the unknown <unk> token for the vocabulary. For \textbf{ResNet Features} extraction, we encode each image with Resnet-101 \cite{resnet} with a 2048-D vector, while we select top $m=50$ salient regions for \textbf{Region Features} with Faster R-CNN \cite{fastercnn}. 
\subsubsection{Module Architecture.}
The AP-Net maps input state $s_t\in\mathbb{R}^{2\times 512}$ into a state embedding $\phi(s_t)\in\mathbb{R}^{512}$ with one fully connected layer and one LeakyReLu layer. The SP-Net takes $\phi(s_t)$ and 512-D embedding for $y_t$ as input, then passes it into a sequence of two fully connected layers with 512 units and 12,186 units. 
\subsubsection{Parameter Settings.}
 The hidden size, all embedding size for images and words are fixed to 512. The batch size for non-attention based models is 32, but 16 for attention-based model.  The learning rate $\mu$ is initiated as $6\times10^{-4}$ then decayed by a factor of 0.8 every three epochs. The discounted factor $\delta$ for discounted imitation learning is set to 0.9. The hyper-parameter $\rho$ and discounted coefficient $\gamma$ are set to 1 and 0.9, respectively. The loss coefficients $\alpha$ and $\beta$ are fixed at 0.2 and 0.8. For compared models, the embedding size of topic vector is set to 100.

\subsection{Comparisons with State-of-The-Art}
\subsubsection{Quantitative Analysis.}
In this section, we quantitatively evaluate various paragraph captioning methods using the standard metrics on the Standford image-paragraph dataset. Here we report the best performance for every model, along with the specification of the language model and the search method at inference stage. \textbf{Greedy} denotes greedy search (equals to $1$-beam search) and \textbf{$n$-beam} indicates the beam search with $n$ most probable sub-sequence \cite{bias}. Generally, more beams used for inference will lead to better performance but higher time-cost. From Table \ref{tab:comparison}, we can observe that our \textbf{CRL} is superior to all the compared paragraph-based and sentence-based image captioning methods in most cases, especially improving CIDEr \cite{cider} by $38.4\%$ (from $22.74\%$ to $31.47\%$).
With only a single layer language decoder, we achieve a significant performance boost over hierarchical methods. Since we select metrics to optimize paragraph-level quality (e.g., CIDEr), the proposed CRL achieves relatively lower performance on the uni-gram metric with synonymous substitution (e.g., METEOR). Regarding observations on compared methods, Non-hierarchical methods (e.g., \textbf{Image-Flat} \cite{NIC}, \textbf{DAM-Att} \cite{DAM} and \textbf{TOMS} \cite{toms}) perform much better than simple concatenation of sentence-level outputs (e.g., \textbf{Neuraltalk} \cite{Karpathy_2015_CVPR}, \textbf{NIC} \cite{NIC}), yet they fail to capture the overall structure and topic transition of paragraphs, thus obtaining a lower performance than hierarchical approaches (e.g., \textbf{Region-Hierarchical} \cite{im2p}, \textbf{RTT-GAN} \cite{RTTGAN} and \textbf{VAE} \cite{eccv}). Different from `Region-Hierarchical' that simply concatenates sentences from the bottom LSTM, `RTT-GAN' and `VAE' preserve a better consistency among sentences. The RL-based method, i.e., \textbf{SCST} \cite{SCST} with single-layer language decoder achieves competitive outcomes compared with the hierarchical model~\cite{im2p}, which demonstrates the power of policy optimization. \textbf{Humans}, as reported in \cite{im2p}, show the results by collecting additional paragraphs for 500 randomly chosen images. We can see that the results show a large gap between automatic synthetic captions and natural language, whereas our proposed CRL with 2-beam search mitigates the gap and achieves competitive outcomes. Besides, experimental results verify that CIDEr metric align better with human judgment than any other evaluation metrics.

\subsubsection{Qualitative Analysis}
In order to intuitively understand the performance of the proposed CRL training, we showcase some outputs with greedy search for randomly selected images in Figure \ref{fig:samples}, \textit{i.e.,} the paragraphs generated by canonical paragraph captioning method `Region-Hierarchical' \cite{im2p}, the proposed CRL method and the RL-based method `SCST' \cite{SCST}. With comparisons with counterparts, our proposed CRL model generates the paragraph in a coherent order: the first sentence (in red) tends to cover a global topic or major actions in visual content, followed by several sentences (in blue) to describe the details of the scene. Generally, the last sentence gives descriptions about objects or environment in the background, which exactly matches human writing styles. Notably, our synthetic paragraphs capture more subtle and accurate words and relationships, such as `platform' and `standing behind the man'. In contrast, both `Region-Hierarchical' and `SCST' could barely guarantee the completeness and richness of the generated paragraphs.

 \begin{figure}[t]
\centering
        \includegraphics[width=0.23\textwidth]{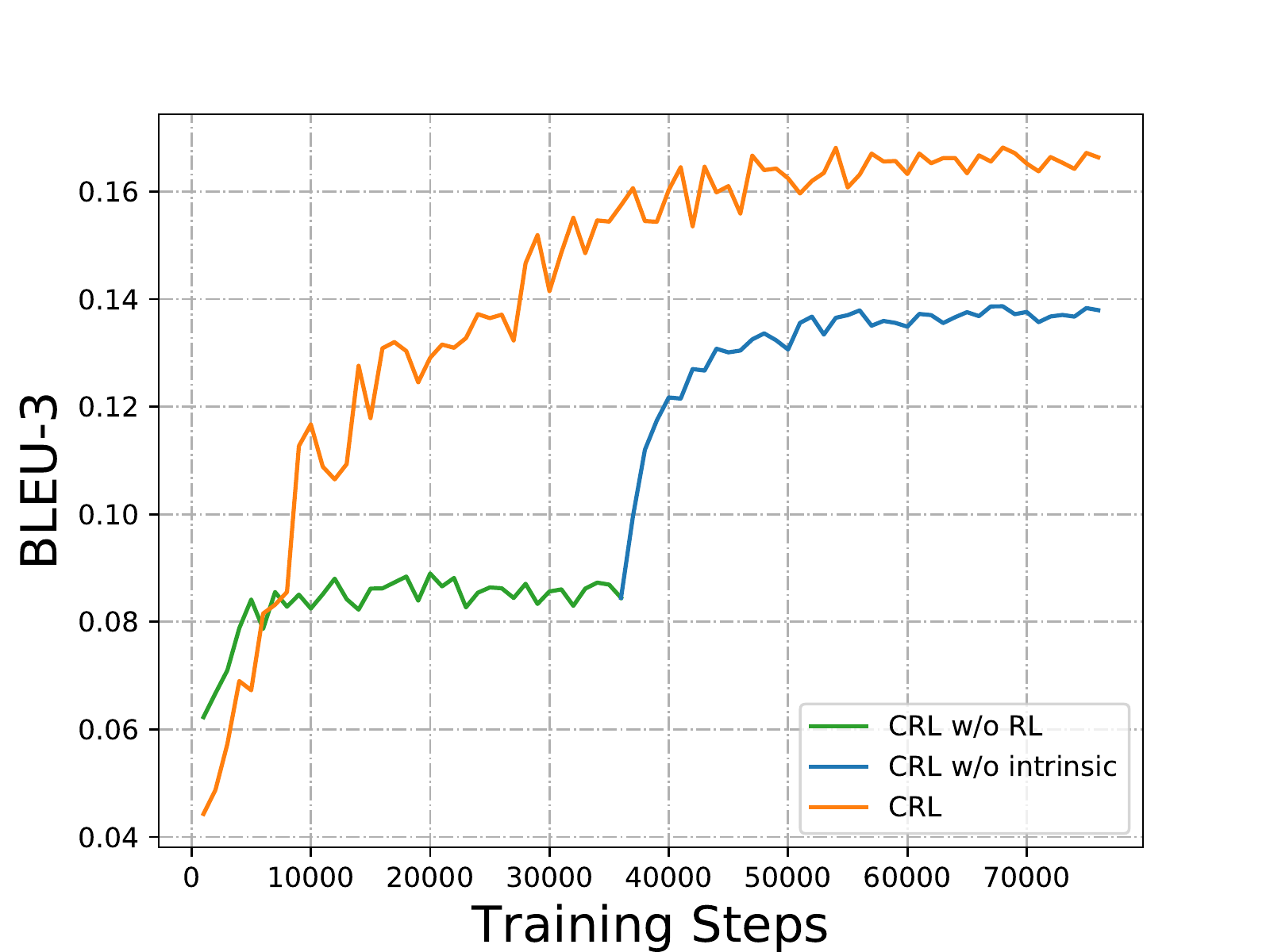}
        \includegraphics[width=0.23\textwidth]{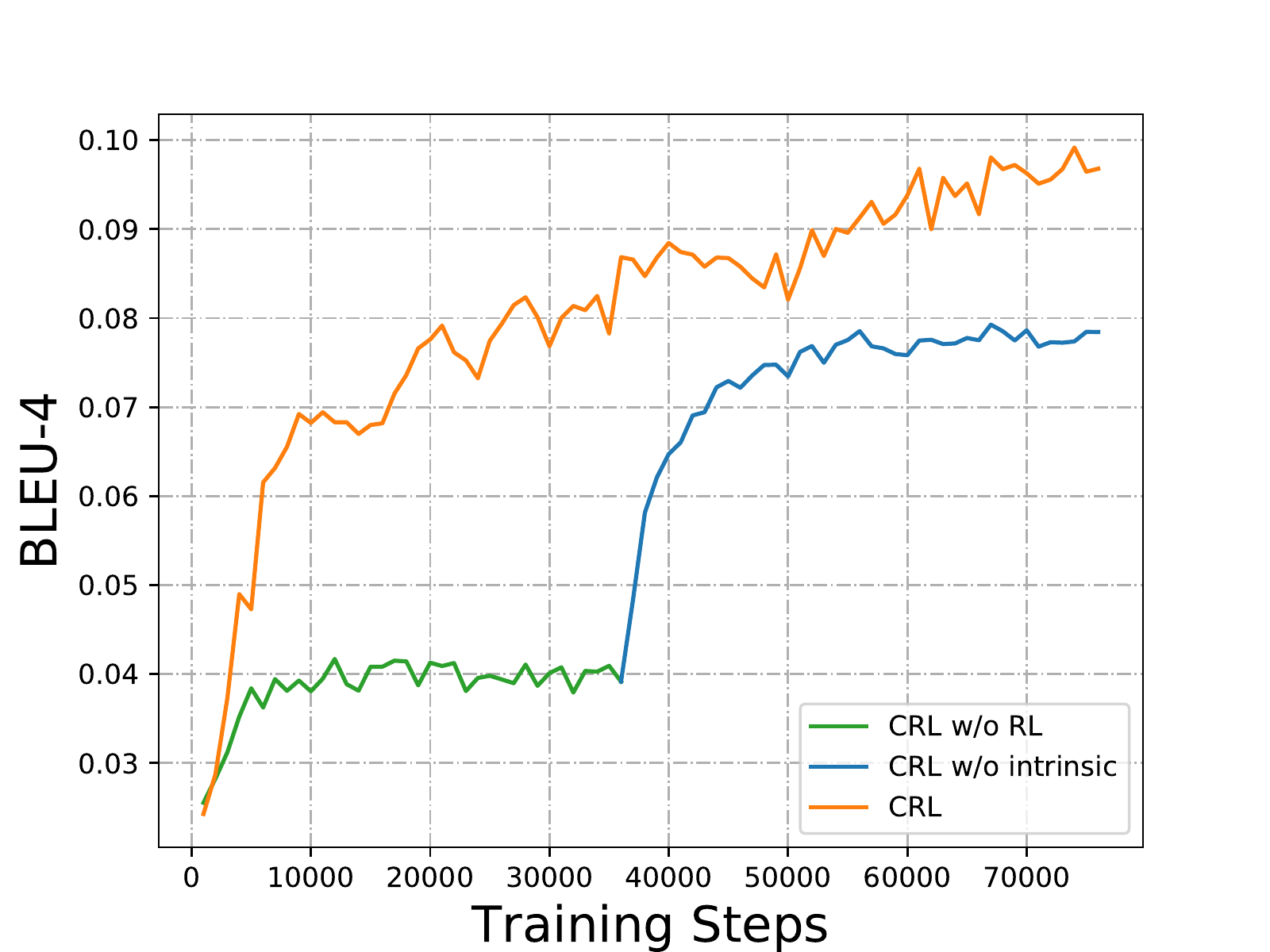}\\
        \includegraphics[width=0.23\textwidth]{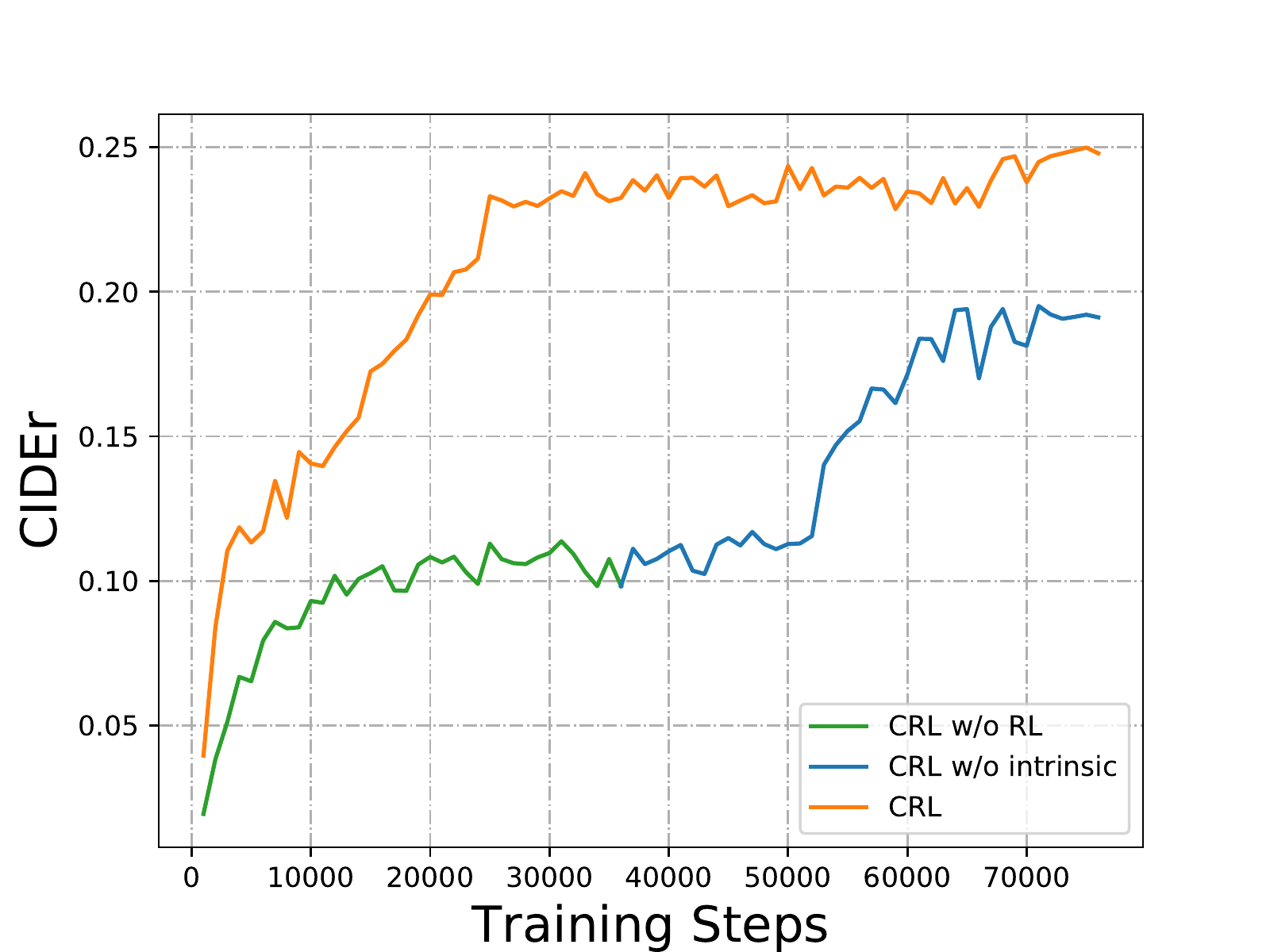}
        \includegraphics[width=0.23\textwidth]{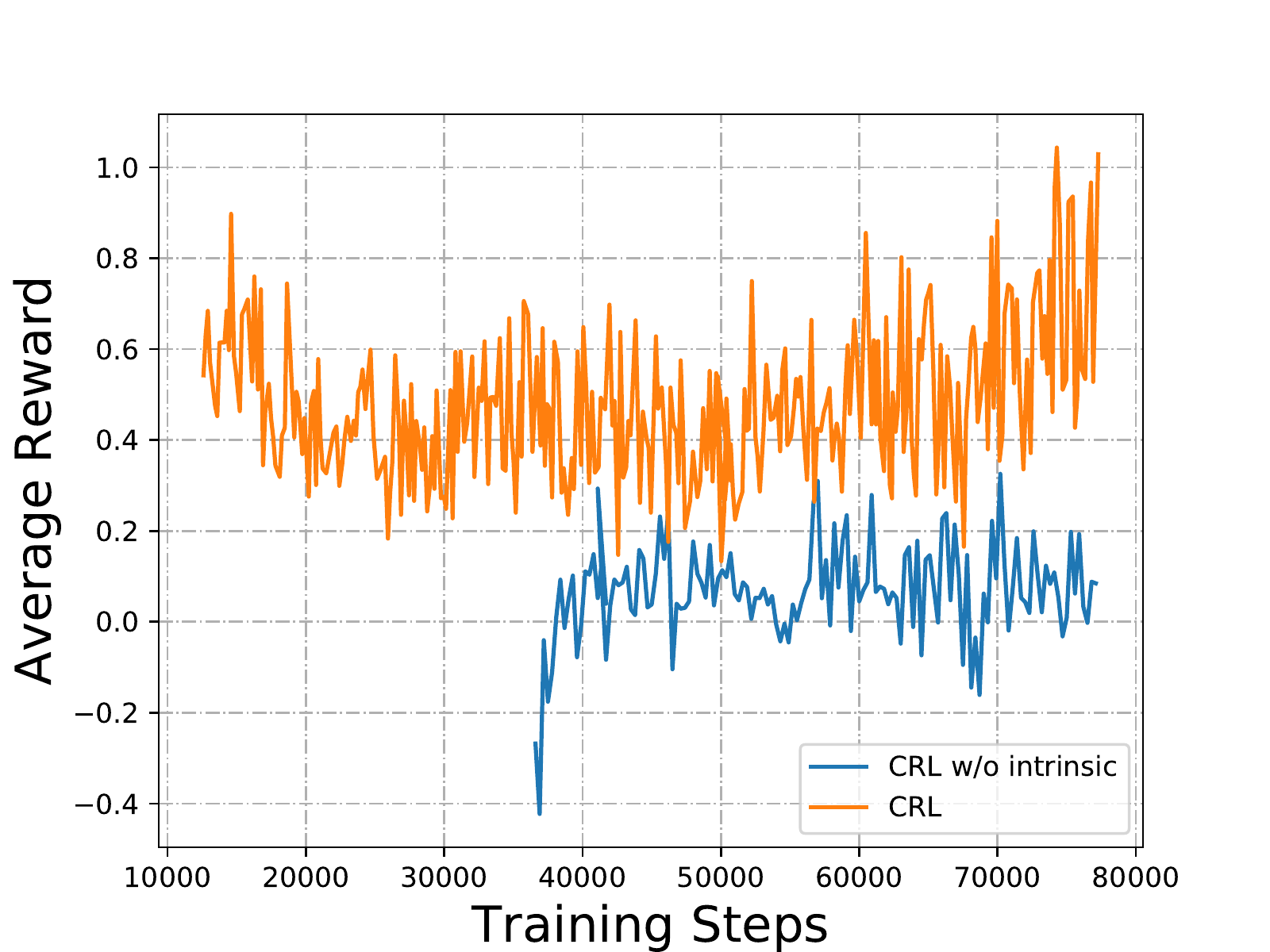}\\
        \vspace{-0.2cm}
\caption{The curves of language measures on validation set and average rewards on Standford image-paragraph dataset.}
\label{fig:curve}
\end{figure}

\begin{figure*}[t]
\centering
        \includegraphics[width=0.73\textwidth]{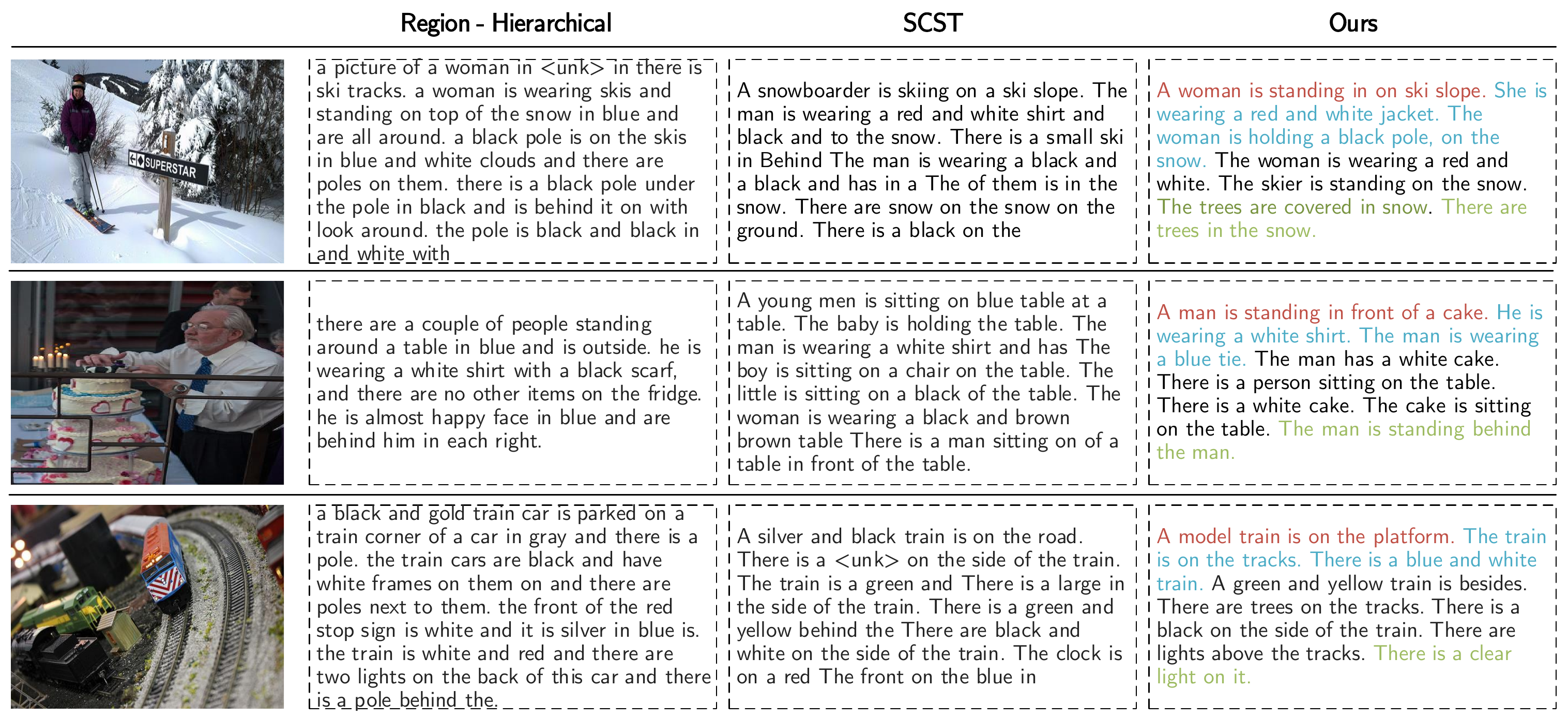}
        \vspace{-0.3cm}
\caption{Paragraphs generated for the images from the Standford image-paragraph dataset. It is observed that our generated paragraphs associated with logic and coherence, \textit{i.e.,} starting with the global sentence marked in red, followed with the details in blue and the background description in green.}
\label{fig:samples}
\end{figure*}

\vspace{-0.3cm}
\begin{figure*}[t]
    \centering
    \transparent{0.8}\includegraphics[width=0.8\linewidth]{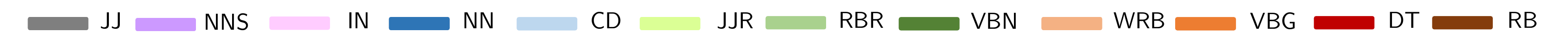}
\end{figure*}
\begin{figure*}[t]
    \centering
    \begin{subfigure}{0.27\linewidth}
        \centering
        \includegraphics[width=0.9\linewidth]{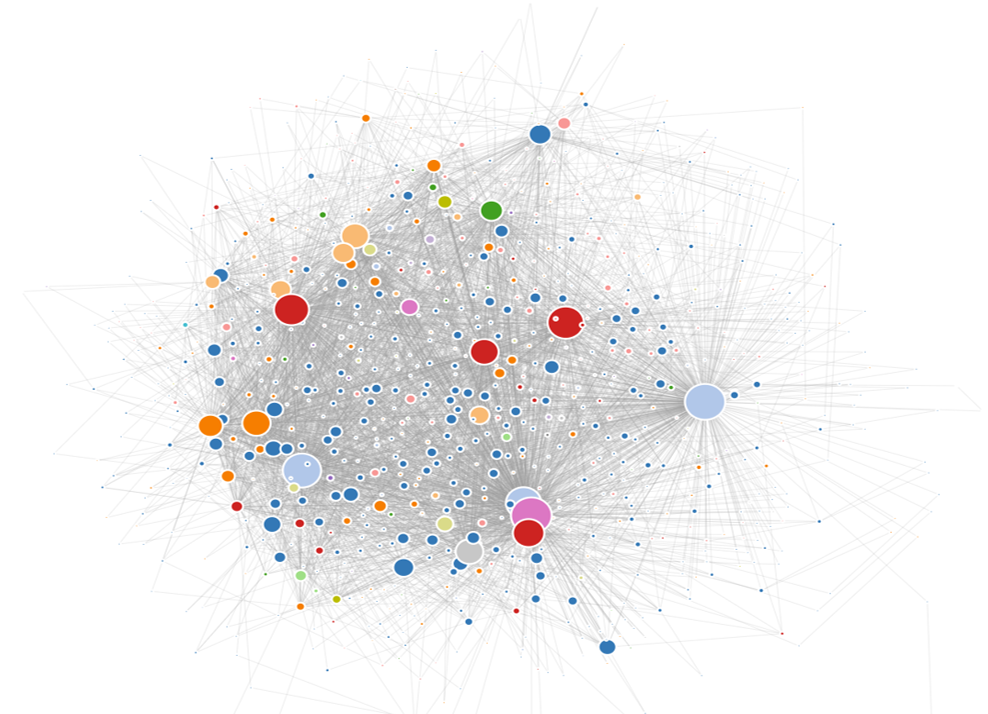}
        \caption{SCST}
    \end{subfigure}
    \begin{subfigure}{0.27\linewidth}
        \centering
        \includegraphics[width=0.9\linewidth]{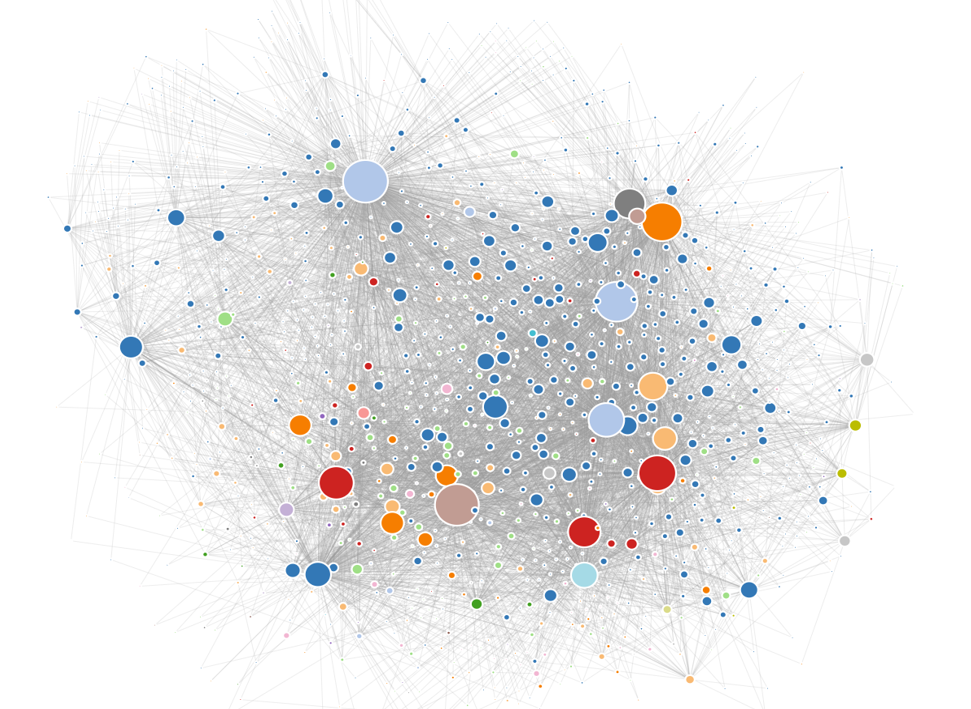}
        \caption{CRL}
    \end{subfigure}
    \begin{subfigure}{0.27\linewidth}
        \centering
        \includegraphics[width=0.9\linewidth]{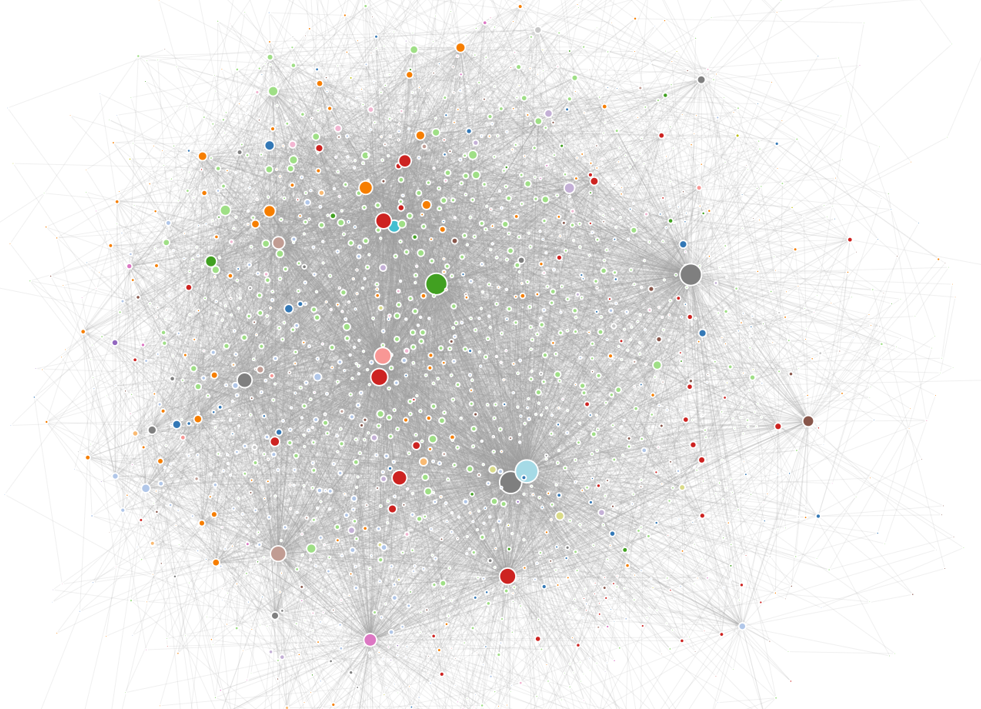}
        \caption{Ground-truth}
    \end{subfigure}%
    \label{fig:graph}
    \vspace{-0.2cm}
\caption{Visualization of diversity of the generated paragraphs by (a) RL-based method SCST \cite{SCST} (b) our proposed CRL method (c) human-beings with semantic network graphs. Each node with different color indicates the unique token with various part-of-speech (POS) tags. Edges show the proximity relationship between tokens.}
\label{fig:graph}
\end{figure*}

\subsection{Ablation Study}
In this section, we study the impact of the RL strategy, the policy network architecture and visual features, respectively. The major experimental results are shown in Table \ref{ablation} and the detailed curve of evaluation metrics and average rewards are illustrated in Figure \ref{fig:curve}.

\subsubsection{RL Training Strategy}
By comparing the performance of each policy used in different methods shown in Table \ref{ablation}, we can observe that pure supervised learning (\textbf{CRL w/o RL}) drops greatly on aggregative metrics like CIDEr and multi-gram metrics like BLEU-\{3,4\}. Removing the intrinsic reward, \textbf{CRL w/o intrinsic} tightly follows the learned policy and lacks essential exploration, thus leading to the sub-optimal performance. Moreover, in Figure \ref{fig:curve}, we show detailed curves of BLEU-$\{3,4\}$, CIDEr per training step and curves of average rewards on validation set based on the `Up-Down' decoder and the `Region Features'. From Figure \ref{fig:curve}, we could observe that `CRL w/o intrinsic' (as the blue line shows) needs a long-period warm-up by `CRL w/o RL' (as the green line shows) and sharply boosts evaluation metrics of the evaluation score after $40$ epochs, and gradually converges afterwards. While our strategy \textbf{CRL} (as the orange line shows) gains a smooth increase during training, since it benefits from the power of combining discounted imitation learning and policy gradient training. Different from `CRL w/o intrinsic', our CRL method avoids tedious and time-consuming initialization, and obtains a full exploration and a better policy network. With respect to the average reward curve, the extrinsic reward achieved by `CRL w/o intrinsic' climbs fast after pre-training, yet our reward signal moves a downhill then slowing uphill. This phenomenon is probably caused by the intrinsic reward, that decreases at very first beginning as the agent learns to control state transition and linguistic patterns. Regarding to variance shown in Figure 4, we infer the variance is mainly introduced by iterative action optimization where the average reward fluctuates correspondingly. The variance can be alleviated by gradient clipping or adjusting learning rate. With the extrinsic rewards that gradually accumulate, the overall reward slowly increases until the model converges. Moreover, it is clearly observed that `CRL' achieves faster convergence (around 25 epochs) compared with `CRL w/o intrinsic' and `CRL w/o RL'.

\subsubsection{Policy Network Architecture}\label{comp}
Regarding to the backbone language decoder, we switch the policy networks from \textbf{FC} (vanilla LSTM \cite{LSTM}), to \textbf{Att} (attention-based LSTM \cite{Att}), to \textbf{Up-Down} (attention-based LSTM + language LSTM \cite{bottomup}) for comparison. 
By comparing each method under different training policies, it is clear that the `Up-Down' model achieves a higher performance among all metrics, as it dynamically attends on local areas of images and captures more visual details. In particular, the `Up-Down' model trained with the proposed curiosity-driven RL averagely increase the CIDEr score by $16.36\%$ and $97.89\%$, the BLEU-4 score by  $14.59\%$ and $98.23\%$ compared with the `CRL w/o instrinsic' and `CRL w/o RL', respectively.

\subsubsection{Visual Features}
In addition to the evaluation considering the impact of visual features, we also list the performance based on \textbf{ResNet Features} and \textbf{Region Features} (see Section \ref{dp} for details). From the Table \ref{ablation}, we can draw the observation that `Region Features' contribute positively to the captioning model, which enriches the visual recognition and representations. Compared with other language networks, the `Up-Down' architecture is more sensitive to the selection of visual features. 

\subsection{Diversity Analysis}
To shed a quantitative light on the linguistic property of generated paragraphs, we randomly select 500 images from the test set of the Stanford image-paragraph dataset, and show the statistics of the paragraphs produced by a representative spread of methods in Figure \ref{fig:graph}. We visualize the semantic graphs of language distribution with a d3-force package in JavaScript. In each semantic graph, each node indicates a unique token from the vocabulary, with different colors to show the associated part-of-speech (POS) tagging. For instance, blue demonstrates Noun in singular form (NN), red shows Determiner (DT) and orange indicates Verb, Gerund or Present Participles (VGB). The edge between two nodes presents the proximity relationship of two words. It is worth noting that the massiveness of semantic graphs imply the diversity and richness of generated paragraphs in an intuitive way. The \textbf{Ground-truth} graph (in Figure 6(c)), annotated by human beings, contains the most comprehensive relationships and extensive object entities. Even though there is still a gap between synthetic paragraphs and real natural language, our generated paragraphs by \textbf{CRL} (in Figure 6(b)) have a much wider vocabulary compared with the one generated by RL-based method \textbf{SCST} (in Figure 6(a)).

\vspace{-0.3cm}
\section{Conclusion}
In this work, we propose an intrinsically motivated reinforcement learning model for visual paragraph generation. Towards generating diverse sentences with coherence, the proposed CRL mines the human writing patterns behind long narratives and well captures precise expressions by modeling the agent's uncertainty of the environment. Distinguishing our work from conventional policy-based and actor-critic based reinforcement learning methods, it alleviates the sparse reward and low exploration issues and thus encourages the agent to fully explore rare states and obtain a better policy.

\eat{Instead of consuming long-term warm-up of policy networks, we naturally integrate the discounted imitation learning into the proposed framework for fast convergence.}
\section{ACKNOWLEDGEMENT}
This work was partially supported by the National Natural Science Foundation of China under Project 61572108 and Project 61632007, Sichuan Science and Technology Program (No. 2018GZDZX0032) and ARC DP 190102353.
\balance{
\bibliographystyle{ACM-Reference-Format}
\bibliography{acmart}


\begin{thebibliography}{56}


\ifx \showCODEN    \undefined \def \showCODEN     #1{\unskip}     \fi
\ifx \showDOI      \undefined \def \showDOI       #1{#1}\fi
\ifx \showISBNx    \undefined \def \showISBNx     #1{\unskip}     \fi
\ifx \showISBNxiii \undefined \def \showISBNxiii  #1{\unskip}     \fi
\ifx \showISSN     \undefined \def \showISSN      #1{\unskip}     \fi
\ifx \showLCCN     \undefined \def \showLCCN      #1{\unskip}     \fi
\ifx \shownote     \undefined \def \shownote      #1{#1}          \fi
\ifx \showarticletitle \undefined \def \showarticletitle #1{#1}   \fi
\ifx \showURL      \undefined \def \showURL       {\relax}        \fi
\providecommand\bibfield[2]{#2}
\providecommand\bibinfo[2]{#2}
\providecommand\natexlab[1]{#1}
\providecommand\showeprint[2][]{arXiv:#2}

\bibitem[\protect\citeauthoryear{Achiam and Sastry}{Achiam and Sastry}{2017}]%
        {surprisal}
\bibfield{author}{\bibinfo{person}{Joshua Achiam} {and}
  \bibinfo{person}{Shankar Sastry}.} \bibinfo{year}{2017}\natexlab{}.
\newblock \showarticletitle{Surprise-Based Intrinsic Motivation for Deep
  Reinforcement Learning}.
\newblock \bibinfo{journal}{\emph{CoRR}}  \bibinfo{volume}{abs/1703.01732}
  (\bibinfo{year}{2017}).
\newblock
\showeprint[arxiv]{1703.01732}
\urldef\tempurl%
\url{http://arxiv.org/abs/1703.01732}
\showURL{%
\tempurl}


\bibitem[\protect\citeauthoryear{Anderson, He, Buehler, Teney, Johnson, Gould,
  and Zhang}{Anderson et~al\mbox{.}}{2018}]%
        {bottomup}
\bibfield{author}{\bibinfo{person}{Peter Anderson}, \bibinfo{person}{Xiaodong
  He}, \bibinfo{person}{Chris Buehler}, \bibinfo{person}{Damien Teney},
  \bibinfo{person}{Mark Johnson}, \bibinfo{person}{Stephen Gould}, {and}
  \bibinfo{person}{Lei Zhang}.} \bibinfo{year}{2018}\natexlab{}.
\newblock \showarticletitle{Bottom-Up and Top-Down Attention for Image
  Captioning and Visual Question Answering}. In \bibinfo{booktitle}{\emph{2018
  {IEEE} Conference on Computer Vision and Pattern Recognition, {CVPR} 2018,
  Salt Lake City, UT, USA, June 18-22, 2018}}. \bibinfo{pages}{6077--6086}.
\newblock


\bibitem[\protect\citeauthoryear{Bellemare, Srinivasan, Ostrovski, Schaul,
  Saxton, and Munos}{Bellemare et~al\mbox{.}}{2016}]%
        {novelty}
\bibfield{author}{\bibinfo{person}{Marc~G. Bellemare}, \bibinfo{person}{Sriram
  Srinivasan}, \bibinfo{person}{Georg Ostrovski}, \bibinfo{person}{Tom Schaul},
  \bibinfo{person}{David Saxton}, {and} \bibinfo{person}{R{\'{e}}mi Munos}.}
  \bibinfo{year}{2016}\natexlab{}.
\newblock \showarticletitle{Unifying Count-Based Exploration and Intrinsic
  Motivation}. In \bibinfo{booktitle}{\emph{Advances in Neural Information
  Processing Systems 29: Annual Conference on Neural Information Processing
  Systems 2016, December 5-10, 2016, Barcelona, Spain}}.
  \bibinfo{pages}{1471--1479}.
\newblock


\bibitem[\protect\citeauthoryear{Bin, Yang, Zhou, Huang, and Shen}{Bin
  et~al\mbox{.}}{2017}]%
        {yang1}
\bibfield{author}{\bibinfo{person}{Yi Bin}, \bibinfo{person}{Yang Yang},
  \bibinfo{person}{Jie Zhou}, \bibinfo{person}{Zi Huang}, {and}
  \bibinfo{person}{Heng~Tao Shen}.} \bibinfo{year}{2017}\natexlab{}.
\newblock \showarticletitle{Adaptively Attending to Visual Attributes and
  Linguistic Knowledge for Captioning}. In
  \bibinfo{booktitle}{\emph{Proceedings of the 2017 {ACM} on Multimedia
  Conference, {MM} 2017, Mountain View, CA, USA, October 23-27, 2017}}.
  \bibinfo{pages}{1345--1353}.
\newblock
\urldef\tempurl%
\url{https://doi.org/10.1145/3123266.3123391}
\showDOI{\tempurl}


\bibitem[\protect\citeauthoryear{Burda, Edwards, Pathak, Storkey, Darrell, and
  Efros}{Burda et~al\mbox{.}}{2018a}]%
        {curiosity1}
\bibfield{author}{\bibinfo{person}{Yuri Burda}, \bibinfo{person}{Harrison
  Edwards}, \bibinfo{person}{Deepak Pathak}, \bibinfo{person}{Amos~J. Storkey},
  \bibinfo{person}{Trevor Darrell}, {and} \bibinfo{person}{Alexei~A. Efros}.}
  \bibinfo{year}{2018}\natexlab{a}.
\newblock \showarticletitle{Large-Scale Study of Curiosity-Driven Learning}.
\newblock \bibinfo{journal}{\emph{CoRR}}  \bibinfo{volume}{abs/1808.04355}
  (\bibinfo{year}{2018}).
\newblock
\showeprint[arxiv]{1808.04355}
\urldef\tempurl%
\url{http://arxiv.org/abs/1808.04355}
\showURL{%
\tempurl}


\bibitem[\protect\citeauthoryear{Burda, Edwards, Storkey, and Klimov}{Burda
  et~al\mbox{.}}{2018b}]%
        {RND}
\bibfield{author}{\bibinfo{person}{Yuri Burda}, \bibinfo{person}{Harrison
  Edwards}, \bibinfo{person}{Amos~J. Storkey}, {and} \bibinfo{person}{Oleg
  Klimov}.} \bibinfo{year}{2018}\natexlab{b}.
\newblock \showarticletitle{Exploration by Random Network Distillation}.
\newblock \bibinfo{journal}{\emph{CoRR}}  \bibinfo{volume}{abs/1810.12894}
  (\bibinfo{year}{2018}).
\newblock
\showeprint[arxiv]{1810.12894}
\urldef\tempurl%
\url{http://arxiv.org/abs/1810.12894}
\showURL{%
\tempurl}


\bibitem[\protect\citeauthoryear{Chatterjee and Schwing}{Chatterjee and
  Schwing}{2018}]%
        {eccv}
\bibfield{author}{\bibinfo{person}{Moitreya Chatterjee} {and}
  \bibinfo{person}{Alexander~G. Schwing}.} \bibinfo{year}{2018}\natexlab{}.
\newblock \showarticletitle{Diverse and Coherent Paragraph Generation from
  Images}. In \bibinfo{booktitle}{\emph{Computer Vision - {ECCV} 2018 - 15th
  European Conference, Munich, Germany, September 8-14, 2018, Proceedings, Part
  {II}}}. \bibinfo{pages}{747--763}.
\newblock


\bibitem[\protect\citeauthoryear{Chen, Ding, Zhao, and Han}{Chen
  et~al\mbox{.}}{2018}]%
        {temporal}
\bibfield{author}{\bibinfo{person}{Hui Chen}, \bibinfo{person}{Guiguang Ding},
  \bibinfo{person}{Sicheng Zhao}, {and} \bibinfo{person}{Jungong Han}.}
  \bibinfo{year}{2018}\natexlab{}.
\newblock \showarticletitle{Temporal-Difference Learning With Sampling Baseline
  for Image Captioning}. In \bibinfo{booktitle}{\emph{Proceedings of the
  Thirty-Second {AAAI} Conference on Artificial Intelligence, (AAAI-18), the
  30th innovative Applications of Artificial Intelligence (IAAI-18), and the
  8th {AAAI} Symposium on Educational Advances in Artificial Intelligence
  (EAAI-18), New Orleans, Louisiana, USA, February 2-7, 2018}}.
  \bibinfo{pages}{6706--6713}.
\newblock


\bibitem[\protect\citeauthoryear{Chen, Liao, Chuang, Hsu, Fu, and Sun}{Chen
  et~al\mbox{.}}{2017}]%
        {adapt}
\bibfield{author}{\bibinfo{person}{Tseng{-}Hung Chen},
  \bibinfo{person}{Yuan{-}Hong Liao}, \bibinfo{person}{Ching{-}Yao Chuang},
  \bibinfo{person}{Wan~Ting Hsu}, \bibinfo{person}{Jianlong Fu}, {and}
  \bibinfo{person}{Min Sun}.} \bibinfo{year}{2017}\natexlab{}.
\newblock \showarticletitle{Show, Adapt and Tell: Adversarial Training of
  Cross-Domain Image Captioner}. In \bibinfo{booktitle}{\emph{{IEEE}
  International Conference on Computer Vision, {ICCV} 2017, Venice, Italy,
  October 22-29, 2017}}. \bibinfo{pages}{521--530}.
\newblock


\bibitem[\protect\citeauthoryear{Dai, Fidler, Urtasun, and Lin}{Dai
  et~al\mbox{.}}{2017}]%
        {conGAN}
\bibfield{author}{\bibinfo{person}{Bo Dai}, \bibinfo{person}{Sanja Fidler},
  \bibinfo{person}{Raquel Urtasun}, {and} \bibinfo{person}{Dahua Lin}.}
  \bibinfo{year}{2017}\natexlab{}.
\newblock \showarticletitle{Towards Diverse and Natural Image Descriptions via
  a Conditional {GAN}}. In \bibinfo{booktitle}{\emph{{IEEE} International
  Conference on Computer Vision, {ICCV} 2017, Venice, Italy, October 22-29,
  2017}}. \bibinfo{pages}{2989--2998}.
\newblock


\bibitem[\protect\citeauthoryear{Denkowski and Lavie}{Denkowski and
  Lavie}{2014}]%
        {meteor}
\bibfield{author}{\bibinfo{person}{Michael~J. Denkowski} {and}
  \bibinfo{person}{Alon Lavie}.} \bibinfo{year}{2014}\natexlab{}.
\newblock \showarticletitle{Meteor Universal: Language Specific Translation
  Evaluation for Any Target Language}. In \bibinfo{booktitle}{\emph{Proceedings
  of the Ninth Workshop on Statistical Machine Translation, WMT@ACL 2014, June
  26-27, 2014, Baltimore, Maryland, {USA}}}. \bibinfo{pages}{376--380}.
\newblock


\bibitem[\protect\citeauthoryear{Goodfellow, Pouget{-}Abadie, Mirza, Xu,
  Warde{-}Farley, Ozair, Courville, and Bengio}{Goodfellow
  et~al\mbox{.}}{2014}]%
        {GAN}
\bibfield{author}{\bibinfo{person}{Ian~J. Goodfellow}, \bibinfo{person}{Jean
  Pouget{-}Abadie}, \bibinfo{person}{Mehdi Mirza}, \bibinfo{person}{Bing Xu},
  \bibinfo{person}{David Warde{-}Farley}, \bibinfo{person}{Sherjil Ozair},
  \bibinfo{person}{Aaron~C. Courville}, {and} \bibinfo{person}{Yoshua Bengio}.}
  \bibinfo{year}{2014}\natexlab{}.
\newblock \showarticletitle{Generative Adversarial Nets}. In
  \bibinfo{booktitle}{\emph{Advances in Neural Information Processing Systems
  27: Annual Conference on Neural Information Processing Systems 2014, December
  8-13 2014, Montreal, Quebec, Canada}}. \bibinfo{pages}{2672--2680}.
\newblock


\bibitem[\protect\citeauthoryear{He, Zhang, Ren, and Sun}{He
  et~al\mbox{.}}{2016}]%
        {resnet}
\bibfield{author}{\bibinfo{person}{Kaiming He}, \bibinfo{person}{Xiangyu
  Zhang}, \bibinfo{person}{Shaoqing Ren}, {and} \bibinfo{person}{Jian Sun}.}
  \bibinfo{year}{2016}\natexlab{}.
\newblock \showarticletitle{Deep Residual Learning for Image Recognition}. In
  \bibinfo{booktitle}{\emph{2016 {IEEE} Conference on Computer Vision and
  Pattern Recognition, {CVPR} 2016, Las Vegas, NV, USA, June 27-30, 2016}}.
  \bibinfo{pages}{770--778}.
\newblock


\bibitem[\protect\citeauthoryear{Hochreiter and Schmidhuber}{Hochreiter and
  Schmidhuber}{1997}]%
        {LSTM}
\bibfield{author}{\bibinfo{person}{Sepp Hochreiter} {and}
  \bibinfo{person}{J{\"{u}}rgen Schmidhuber}.} \bibinfo{year}{1997}\natexlab{}.
\newblock \showarticletitle{Long Short-Term Memory}.
\newblock \bibinfo{journal}{\emph{Neural Computation}} \bibinfo{volume}{9},
  \bibinfo{number}{8} (\bibinfo{year}{1997}), \bibinfo{pages}{1735--1780}.
\newblock


\bibitem[\protect\citeauthoryear{Houthooft, Chen, Duan, Schulman, Turck, and
  Abbeel}{Houthooft et~al\mbox{.}}{2016}]%
        {novelty3}
\bibfield{author}{\bibinfo{person}{Rein Houthooft}, \bibinfo{person}{Xi Chen},
  \bibinfo{person}{Yan Duan}, \bibinfo{person}{John Schulman},
  \bibinfo{person}{Filip~De Turck}, {and} \bibinfo{person}{Pieter Abbeel}.}
  \bibinfo{year}{2016}\natexlab{}.
\newblock \showarticletitle{{VIME:} Variational Information Maximizing
  Exploration}. In \bibinfo{booktitle}{\emph{Advances in Neural Information
  Processing Systems 29: Annual Conference on Neural Information Processing
  Systems 2016, December 5-10, 2016, Barcelona, Spain}}.
  \bibinfo{pages}{1109--1117}.
\newblock


\bibitem[\protect\citeauthoryear{Huang, Ferraro, Mostafazadeh, Misra, Agrawal,
  Devlin, Girshick, He, Kohli, Batra, Zitnick, Parikh, Vanderwende, Galley, and
  Mitchell}{Huang et~al\mbox{.}}{2016}]%
        {visualstory}
\bibfield{author}{\bibinfo{person}{Ting{-}Hao~(Kenneth) Huang},
  \bibinfo{person}{Francis Ferraro}, \bibinfo{person}{Nasrin Mostafazadeh},
  \bibinfo{person}{Ishan Misra}, \bibinfo{person}{Aishwarya Agrawal},
  \bibinfo{person}{Jacob Devlin}, \bibinfo{person}{Ross~B. Girshick},
  \bibinfo{person}{Xiaodong He}, \bibinfo{person}{Pushmeet Kohli},
  \bibinfo{person}{Dhruv Batra}, \bibinfo{person}{C.~Lawrence Zitnick},
  \bibinfo{person}{Devi Parikh}, \bibinfo{person}{Lucy Vanderwende},
  \bibinfo{person}{Michel Galley}, {and} \bibinfo{person}{Margaret Mitchell}.}
  \bibinfo{year}{2016}\natexlab{}.
\newblock \showarticletitle{Visual Storytelling}. In
  \bibinfo{booktitle}{\emph{{NAACL} {HLT} 2016, The 2016 Conference of the
  North American Chapter of the Association for Computational Linguistics:
  Human Language Technologies, San Diego California, USA, June 12-17, 2016}}.
  \bibinfo{pages}{1233--1239}.
\newblock


\bibitem[\protect\citeauthoryear{Johnson, Karpathy, and Fei{-}Fei}{Johnson
  et~al\mbox{.}}{2016}]%
        {densecap}
\bibfield{author}{\bibinfo{person}{Justin Johnson}, \bibinfo{person}{Andrej
  Karpathy}, {and} \bibinfo{person}{Li Fei{-}Fei}.}
  \bibinfo{year}{2016}\natexlab{}.
\newblock \showarticletitle{DenseCap: Fully Convolutional Localization Networks
  for Dense Captioning}. In \bibinfo{booktitle}{\emph{2016 {IEEE} Conference on
  Computer Vision and Pattern Recognition, {CVPR} 2016, Las Vegas, NV, USA,
  June 27-30, 2016}}. \bibinfo{pages}{4565--4574}.
\newblock


\bibitem[\protect\citeauthoryear{Karpathy and Li}{Karpathy and Li}{2015}]%
        {Karpathy_2015_CVPR}
\bibfield{author}{\bibinfo{person}{Andrej Karpathy} {and}
  \bibinfo{person}{Fei{-}Fei Li}.} \bibinfo{year}{2015}\natexlab{}.
\newblock \showarticletitle{Deep visual-semantic alignments for generating
  image descriptions}. In \bibinfo{booktitle}{\emph{{IEEE} Conference on
  Computer Vision and Pattern Recognition, {CVPR} 2015, Boston, MA, USA, June
  7-12, 2015}}. \bibinfo{pages}{3128--3137}.
\newblock


\bibitem[\protect\citeauthoryear{Kingma and Ba}{Kingma and Ba}{2015}]%
        {adam}
\bibfield{author}{\bibinfo{person}{Diederik~P. Kingma} {and}
  \bibinfo{person}{Jimmy Ba}.} \bibinfo{year}{2015}\natexlab{}.
\newblock \showarticletitle{Adam: {A} Method for Stochastic Optimization}. In
  \bibinfo{booktitle}{\emph{3rd International Conference on Learning
  Representations, {ICLR} 2015, San Diego, CA, USA, May 7-9, 2015, Conference
  Track Proceedings}}.
\newblock


\bibitem[\protect\citeauthoryear{Kingma and Welling}{Kingma and
  Welling}{2013}]%
        {VAE}
\bibfield{author}{\bibinfo{person}{Diederik~P. Kingma} {and}
  \bibinfo{person}{Max Welling}.} \bibinfo{year}{2013}\natexlab{}.
\newblock \showarticletitle{Auto-Encoding Variational Bayes}.
\newblock \bibinfo{journal}{\emph{CoRR}}  \bibinfo{volume}{abs/1312.6114}
  (\bibinfo{year}{2013}).
\newblock
\showeprint[arxiv]{1312.6114}
\urldef\tempurl%
\url{http://arxiv.org/abs/1312.6114}
\showURL{%
\tempurl}


\bibitem[\protect\citeauthoryear{Krause, Johnson, Krishna, and
  Fei{-}Fei}{Krause et~al\mbox{.}}{2017}]%
        {im2p}
\bibfield{author}{\bibinfo{person}{Jonathan Krause}, \bibinfo{person}{Justin
  Johnson}, \bibinfo{person}{Ranjay Krishna}, {and} \bibinfo{person}{Li
  Fei{-}Fei}.} \bibinfo{year}{2017}\natexlab{}.
\newblock \showarticletitle{A Hierarchical Approach for Generating Descriptive
  Image Paragraphs}. In \bibinfo{booktitle}{\emph{2017 {IEEE} Conference on
  Computer Vision and Pattern Recognition, {CVPR} 2017, Honolulu, HI, USA, July
  21-26, 2017}}. \bibinfo{pages}{3337--3345}.
\newblock


\bibitem[\protect\citeauthoryear{Li, Luong, and Jurafsky}{Li
  et~al\mbox{.}}{2015}]%
        {para}
\bibfield{author}{\bibinfo{person}{Jiwei Li}, \bibinfo{person}{Minh{-}Thang
  Luong}, {and} \bibinfo{person}{Dan Jurafsky}.}
  \bibinfo{year}{2015}\natexlab{}.
\newblock \showarticletitle{A Hierarchical Neural Autoencoder for Paragraphs
  and Documents}. In \bibinfo{booktitle}{\emph{Proceedings of the 53rd Annual
  Meeting of the Association for Computational Linguistics and the 7th
  International Joint Conference on Natural Language Processing of the Asian
  Federation of Natural Language Processing, {ACL} 2015, July 26-31, 2015,
  Beijing, China, Volume 1: Long Papers}}. \bibinfo{pages}{1106--1115}.
\newblock


\bibitem[\protect\citeauthoryear{Li, Luo, Zhang, Sadiq, and Cui}{Li
  et~al\mbox{.}}{2019}]%
        {yadan1}
\bibfield{author}{\bibinfo{person}{Yang Li}, \bibinfo{person}{Yadan Luo},
  \bibinfo{person}{Zheng Zhang}, \bibinfo{person}{Shazia Sadiq}, {and}
  \bibinfo{person}{Peng Cui}.} \bibinfo{year}{2019}\natexlab{}.
\newblock \showarticletitle{Context-Aware Attention-Based Data Augmentation for
  {POI} Recommendation}. In \bibinfo{booktitle}{\emph{35th {IEEE} International
  Conference on Data Engineering Workshops, {ICDE} Workshops 2019, Macao,
  China, April 8-12, 2019}}. \bibinfo{pages}{177--184}.
\newblock


\bibitem[\protect\citeauthoryear{Liang, Hu, Zhang, Gan, and Xing}{Liang
  et~al\mbox{.}}{2017}]%
        {RTTGAN}
\bibfield{author}{\bibinfo{person}{Xiaodan Liang}, \bibinfo{person}{Zhiting
  Hu}, \bibinfo{person}{Hao Zhang}, \bibinfo{person}{Chuang Gan}, {and}
  \bibinfo{person}{Eric~P. Xing}.} \bibinfo{year}{2017}\natexlab{}.
\newblock \showarticletitle{Recurrent Topic-Transition {GAN} for Visual
  Paragraph Generation}. In \bibinfo{booktitle}{\emph{{IEEE} International
  Conference on Computer Vision, {ICCV} 2017, Venice, Italy, October 22-29,
  2017}}. \bibinfo{pages}{3382--3391}.
\newblock


\bibitem[\protect\citeauthoryear{Liu, Zha, Zhang, Zhang, and Wu}{Liu
  et~al\mbox{.}}{2018c}]%
        {hanwang}
\bibfield{author}{\bibinfo{person}{Daqing Liu}, \bibinfo{person}{Zheng{-}Jun
  Zha}, \bibinfo{person}{Hanwang Zhang}, \bibinfo{person}{Yongdong Zhang},
  {and} \bibinfo{person}{Feng Wu}.} \bibinfo{year}{2018}\natexlab{c}.
\newblock \showarticletitle{Context-Aware Visual Policy Network for
  Sequence-Level Image Captioning}. In \bibinfo{booktitle}{\emph{2018 {ACM}
  Multimedia Conference on Multimedia Conference, {MM} 2018, Seoul, Republic of
  Korea, October 22-26, 2018}}. \bibinfo{pages}{1416--1424}.
\newblock


\bibitem[\protect\citeauthoryear{Liu, Wan, and Guo}{Liu et~al\mbox{.}}{2018a}]%
        {poetry}
\bibfield{author}{\bibinfo{person}{Lixin Liu}, \bibinfo{person}{Xiaojun Wan},
  {and} \bibinfo{person}{Zongming Guo}.} \bibinfo{year}{2018}\natexlab{a}.
\newblock \showarticletitle{Images2Poem: Generating Chinese Poetry from Image
  Streams}. In \bibinfo{booktitle}{\emph{2018 {ACM} Multimedia Conference on
  Multimedia Conference, {MM} 2018, Seoul, Republic of Korea, October 22-26,
  2018}}. \bibinfo{pages}{1967--1975}.
\newblock


\bibitem[\protect\citeauthoryear{Liu, Wan, and Guo}{Liu et~al\mbox{.}}{2018b}]%
        {poetry1}
\bibfield{author}{\bibinfo{person}{Lixin Liu}, \bibinfo{person}{Xiaojun Wan},
  {and} \bibinfo{person}{Zongming Guo}.} \bibinfo{year}{2018}\natexlab{b}.
\newblock \showarticletitle{Images2Poem: Generating Chinese Poetry from Image
  Streams}. In \bibinfo{booktitle}{\emph{2018 {ACM} Multimedia Conference on
  Multimedia Conference, {MM} 2018, Seoul, Republic of Korea, October 22-26,
  2018}}. \bibinfo{pages}{1967--1975}.
\newblock


\bibitem[\protect\citeauthoryear{Liu, Zhu, Ye, Guadarrama, and Murphy}{Liu
  et~al\mbox{.}}{2017}]%
        {SPIDEr}
\bibfield{author}{\bibinfo{person}{Siqi Liu}, \bibinfo{person}{Zhenhai Zhu},
  \bibinfo{person}{Ning Ye}, \bibinfo{person}{Sergio Guadarrama}, {and}
  \bibinfo{person}{Kevin Murphy}.} \bibinfo{year}{2017}\natexlab{}.
\newblock \showarticletitle{Improved Image Captioning via Policy Gradient
  optimization of SPIDEr}. In \bibinfo{booktitle}{\emph{{IEEE} International
  Conference on Computer Vision, {ICCV} 2017, Venice, Italy, October 22-29,
  2017}}. \bibinfo{pages}{873--881}.
\newblock


\bibitem[\protect\citeauthoryear{Lopes, Lang, Toussaint, and Oudeyer}{Lopes
  et~al\mbox{.}}{2012}]%
        {novelty1}
\bibfield{author}{\bibinfo{person}{Manuel Lopes}, \bibinfo{person}{Tobias
  Lang}, \bibinfo{person}{Marc Toussaint}, {and} \bibinfo{person}{Pierre{-}Yves
  Oudeyer}.} \bibinfo{year}{2012}\natexlab{}.
\newblock \showarticletitle{Exploration in Model-based Reinforcement Learning
  by Empirically Estimating Learning Progress}. In
  \bibinfo{booktitle}{\emph{Advances in Neural Information Processing Systems
  25: 26th Annual Conference on Neural Information Processing Systems 2012.
  Proceedings of a meeting held December 3-6, 2012, Lake Tahoe, Nevada, United
  States.}} \bibinfo{pages}{206--214}.
\newblock


\bibitem[\protect\citeauthoryear{Luo, Wang, Huang, Yang, and Zhao}{Luo
  et~al\mbox{.}}{2018}]%
        {yadan}
\bibfield{author}{\bibinfo{person}{Yadan Luo}, \bibinfo{person}{Ziwei Wang},
  \bibinfo{person}{Zi Huang}, \bibinfo{person}{Yang Yang}, {and}
  \bibinfo{person}{Cong Zhao}.} \bibinfo{year}{2018}\natexlab{}.
\newblock \showarticletitle{Coarse-to-Fine Annotation Enrichment for Semantic
  Segmentation Learning}. In \bibinfo{booktitle}{\emph{Proceedings of the 27th
  {ACM} International Conference on Information and Knowledge Management,
  {CIKM} 2018, Torino, Italy, October 22-26, 2018}}. \bibinfo{pages}{237--246}.
\newblock


\bibitem[\protect\citeauthoryear{Mao, Zhou, Wang, and Li}{Mao
  et~al\mbox{.}}{2018}]%
        {toms}
\bibfield{author}{\bibinfo{person}{Yuzhao Mao}, \bibinfo{person}{Chang Zhou},
  \bibinfo{person}{Xiaojie Wang}, {and} \bibinfo{person}{Ruifan Li}.}
  \bibinfo{year}{2018}\natexlab{}.
\newblock \showarticletitle{Show and Tell More: Topic-Oriented Multi-Sentence
  Image Captioning}. In \bibinfo{booktitle}{\emph{Proceedings of the
  Twenty-Seventh International Joint Conference on Artificial Intelligence,
  {IJCAI} 2018, July 13-19, 2018, Stockholm, Sweden.}}
  \bibinfo{pages}{4258--4264}.
\newblock


\bibitem[\protect\citeauthoryear{Papineni, Roukos, Ward, and Zhu}{Papineni
  et~al\mbox{.}}{2002}]%
        {bleu}
\bibfield{author}{\bibinfo{person}{Kishore Papineni}, \bibinfo{person}{Salim
  Roukos}, \bibinfo{person}{Todd Ward}, {and} \bibinfo{person}{Wei{-}Jing
  Zhu}.} \bibinfo{year}{2002}\natexlab{}.
\newblock \showarticletitle{Bleu: a Method for Automatic Evaluation of Machine
  Translation}. In \bibinfo{booktitle}{\emph{Proceedings of the 40th Annual
  Meeting of the Association for Computational Linguistics, July 6-12, 2002,
  Philadelphia, PA, {USA.}}} \bibinfo{pages}{311--318}.
\newblock


\bibitem[\protect\citeauthoryear{Paszke, Gross, Chintala, Chanan, Yang, DeVito,
  Lin, Desmaison, Antiga, and Lerer}{Paszke et~al\mbox{.}}{2017}]%
        {pytorch}
\bibfield{author}{\bibinfo{person}{Adam Paszke}, \bibinfo{person}{Sam Gross},
  \bibinfo{person}{Soumith Chintala}, \bibinfo{person}{Gregory Chanan},
  \bibinfo{person}{Edward Yang}, \bibinfo{person}{Zachary DeVito},
  \bibinfo{person}{Zeming Lin}, \bibinfo{person}{Alban Desmaison},
  \bibinfo{person}{Luca Antiga}, {and} \bibinfo{person}{Adam Lerer}.}
  \bibinfo{year}{2017}\natexlab{}.
\newblock \showarticletitle{Automatic differentiation in PyTorch}.
\newblock  (\bibinfo{year}{2017}).
\newblock


\bibitem[\protect\citeauthoryear{Pathak, Agrawal, Efros, and Darrell}{Pathak
  et~al\mbox{.}}{2017}]%
        {curiosity}
\bibfield{author}{\bibinfo{person}{Deepak Pathak}, \bibinfo{person}{Pulkit
  Agrawal}, \bibinfo{person}{Alexei~A. Efros}, {and} \bibinfo{person}{Trevor
  Darrell}.} \bibinfo{year}{2017}\natexlab{}.
\newblock \showarticletitle{Curiosity-driven Exploration by Self-supervised
  Prediction}. In \bibinfo{booktitle}{\emph{Proceedings of the 34th
  International Conference on Machine Learning, {ICML} 2017, Sydney, NSW,
  Australia, 6-11 August 2017}}. \bibinfo{pages}{2778--2787}.
\newblock


\bibitem[\protect\citeauthoryear{Ranzato, Chopra, Auli, and Zaremba}{Ranzato
  et~al\mbox{.}}{2016}]%
        {bias}
\bibfield{author}{\bibinfo{person}{Marc'Aurelio Ranzato},
  \bibinfo{person}{Sumit Chopra}, \bibinfo{person}{Michael Auli}, {and}
  \bibinfo{person}{Wojciech Zaremba}.} \bibinfo{year}{2016}\natexlab{}.
\newblock \showarticletitle{Sequence Level Training with Recurrent Neural
  Networks}. In \bibinfo{booktitle}{\emph{{ICLR}}}.
\newblock
\urldef\tempurl%
\url{http://arxiv.org/abs/1511.06732}
\showURL{%
\tempurl}


\bibitem[\protect\citeauthoryear{Ravi, Wang, Mu{\~{n}}iz, Sigal, Metaxas, and
  Kapadia}{Ravi et~al\mbox{.}}{2018}]%
        {story2}
\bibfield{author}{\bibinfo{person}{Hareesh Ravi}, \bibinfo{person}{Lezi Wang},
  \bibinfo{person}{Carlos Mu{\~{n}}iz}, \bibinfo{person}{Leonid Sigal},
  \bibinfo{person}{Dimitris~N. Metaxas}, {and} \bibinfo{person}{Mubbasir
  Kapadia}.} \bibinfo{year}{2018}\natexlab{}.
\newblock \showarticletitle{Show Me a Story: Towards Coherent Neural Story
  Illustration}. In \bibinfo{booktitle}{\emph{2018 {IEEE} Conference on
  Computer Vision and Pattern Recognition, {CVPR} 2018, Salt Lake City, UT,
  USA, June 18-22, 2018}}. \bibinfo{pages}{7613--7621}.
\newblock


\bibitem[\protect\citeauthoryear{Ren, He, Girshick, and Sun}{Ren
  et~al\mbox{.}}{2015}]%
        {fastercnn}
\bibfield{author}{\bibinfo{person}{Shaoqing Ren}, \bibinfo{person}{Kaiming He},
  \bibinfo{person}{Ross~B. Girshick}, {and} \bibinfo{person}{Jian Sun}.}
  \bibinfo{year}{2015}\natexlab{}.
\newblock \showarticletitle{Faster {R-CNN:} Towards Real-Time Object Detection
  with Region Proposal Networks}. In \bibinfo{booktitle}{\emph{Advances in
  Neural Information Processing Systems 28: Annual Conference on Neural
  Information Processing Systems 2015, December 7-12, 2015, Montreal, Quebec,
  Canada}}. \bibinfo{pages}{91--99}.
\newblock


\bibitem[\protect\citeauthoryear{Ren, Wang, Zhang, Lv, and Li}{Ren
  et~al\mbox{.}}{2017}]%
        {ER}
\bibfield{author}{\bibinfo{person}{Zhou Ren}, \bibinfo{person}{Xiaoyu Wang},
  \bibinfo{person}{Ning Zhang}, \bibinfo{person}{Xutao Lv}, {and}
  \bibinfo{person}{Li{-}Jia Li}.} \bibinfo{year}{2017}\natexlab{}.
\newblock \showarticletitle{Deep Reinforcement Learning-Based Image Captioning
  with Embedding Reward}. In \bibinfo{booktitle}{\emph{2017 {IEEE} Conference
  on Computer Vision and Pattern Recognition, {CVPR} 2017, Honolulu, HI, USA,
  July 21-26, 2017}}. \bibinfo{pages}{1151--1159}.
\newblock


\bibitem[\protect\citeauthoryear{Rennie, Marcheret, Mroueh, Ross, and
  Goel}{Rennie et~al\mbox{.}}{2017}]%
        {SCST}
\bibfield{author}{\bibinfo{person}{Steven~J. Rennie}, \bibinfo{person}{Etienne
  Marcheret}, \bibinfo{person}{Youssef Mroueh}, \bibinfo{person}{Jarret Ross},
  {and} \bibinfo{person}{Vaibhava Goel}.} \bibinfo{year}{2017}\natexlab{}.
\newblock \showarticletitle{Self-Critical Sequence Training for Image
  Captioning}. In \bibinfo{booktitle}{\emph{2017 {IEEE} Conference on Computer
  Vision and Pattern Recognition, {CVPR} 2017, Honolulu, HI, USA, July 21-26,
  2017}}. \bibinfo{pages}{1179--1195}.
\newblock


\bibitem[\protect\citeauthoryear{Singh, Barto, and Chentanez}{Singh
  et~al\mbox{.}}{2004}]%
        {intrinsic}
\bibfield{author}{\bibinfo{person}{Satinder~P. Singh},
  \bibinfo{person}{Andrew~G. Barto}, {and} \bibinfo{person}{Nuttapong
  Chentanez}.} \bibinfo{year}{2004}\natexlab{}.
\newblock \showarticletitle{Intrinsically Motivated Reinforcement Learning}. In
  \bibinfo{booktitle}{\emph{Advances in Neural Information Processing Systems
  17 [Neural Information Processing Systems, {NIPS} 2004, December 13-18, 2004,
  Vancouver, British Columbia, Canada]}}. \bibinfo{pages}{1281--1288}.
\newblock


\bibitem[\protect\citeauthoryear{Sutton, Barto, Bach, et~al\mbox{.}}{Sutton
  et~al\mbox{.}}{2017}]%
        {sutton}
\bibfield{author}{\bibinfo{person}{Richard~S Sutton}, \bibinfo{person}{Andrew~G
  Barto}, \bibinfo{person}{Francis Bach}, {et~al\mbox{.}}}
  \bibinfo{year}{2017}\natexlab{}.
\newblock \bibinfo{booktitle}{\emph{Reinforcement learning: An introduction
  (2nd Edition)}}.
\newblock \bibinfo{publisher}{MIT press}.
\newblock


\bibitem[\protect\citeauthoryear{Tang, Houthooft, Foote, Stooke, Chen, Duan,
  Schulman, Turck, and Abbeel}{Tang et~al\mbox{.}}{2017}]%
        {novelty2}
\bibfield{author}{\bibinfo{person}{Haoran Tang}, \bibinfo{person}{Rein
  Houthooft}, \bibinfo{person}{Davis Foote}, \bibinfo{person}{Adam Stooke},
  \bibinfo{person}{Xi Chen}, \bibinfo{person}{Yan Duan}, \bibinfo{person}{John
  Schulman}, \bibinfo{person}{Filip~De Turck}, {and} \bibinfo{person}{Pieter
  Abbeel}.} \bibinfo{year}{2017}\natexlab{}.
\newblock \showarticletitle{{\#}Exploration: {A} Study of Count-Based
  Exploration for Deep Reinforcement Learning}. In
  \bibinfo{booktitle}{\emph{Advances in Neural Information Processing Systems
  30: Annual Conference on Neural Information Processing Systems 2017, 4-9
  December 2017, Long Beach, CA, {USA}}}. \bibinfo{pages}{2750--2759}.
\newblock


\bibitem[\protect\citeauthoryear{Vedantam, Zitnick, and Parikh}{Vedantam
  et~al\mbox{.}}{2015}]%
        {cider}
\bibfield{author}{\bibinfo{person}{Ramakrishna Vedantam},
  \bibinfo{person}{C.~Lawrence Zitnick}, {and} \bibinfo{person}{Devi Parikh}.}
  \bibinfo{year}{2015}\natexlab{}.
\newblock \showarticletitle{CIDEr: Consensus-based image description
  evaluation}. In \bibinfo{booktitle}{\emph{{IEEE} Conference on Computer
  Vision and Pattern Recognition, {CVPR} 2015, Boston, MA, USA, June 7-12,
  2015}}. \bibinfo{pages}{4566--4575}.
\newblock


\bibitem[\protect\citeauthoryear{Vinyals, Toshev, Bengio, and Erhan}{Vinyals
  et~al\mbox{.}}{2015}]%
        {NIC}
\bibfield{author}{\bibinfo{person}{Oriol Vinyals}, \bibinfo{person}{Alexander
  Toshev}, \bibinfo{person}{Samy Bengio}, {and} \bibinfo{person}{Dumitru
  Erhan}.} \bibinfo{year}{2015}\natexlab{}.
\newblock \showarticletitle{Show and tell: {A} neural image caption generator}.
  In \bibinfo{booktitle}{\emph{{IEEE} Conference on Computer Vision and Pattern
  Recognition, {CVPR} 2015, Boston, MA, USA, June 7-12, 2015}}.
  \bibinfo{pages}{3156--3164}.
\newblock


\bibitem[\protect\citeauthoryear{Wang, Chen, Wang, and Wang}{Wang
  et~al\mbox{.}}{2018a}]%
        {visualstory1}
\bibfield{author}{\bibinfo{person}{Xin Wang}, \bibinfo{person}{Wenhu Chen},
  \bibinfo{person}{Yuan{-}Fang Wang}, {and} \bibinfo{person}{William~Yang
  Wang}.} \bibinfo{year}{2018}\natexlab{a}.
\newblock \showarticletitle{No Metrics Are Perfect: Adversarial Reward Learning
  for Visual Storytelling}. In \bibinfo{booktitle}{\emph{Proceedings of the
  56th Annual Meeting of the Association for Computational Linguistics, {ACL}
  2018, Melbourne, Australia, July 15-20, 2018, Volume 1: Long Papers}}.
  \bibinfo{pages}{899--909}.
\newblock


\bibitem[\protect\citeauthoryear{Wang, Chen, Wu, Wang, and Wang}{Wang
  et~al\mbox{.}}{2018b}]%
        {HRL}
\bibfield{author}{\bibinfo{person}{Xin Wang}, \bibinfo{person}{Wenhu Chen},
  \bibinfo{person}{Jiawei Wu}, \bibinfo{person}{Yuan{-}Fang Wang}, {and}
  \bibinfo{person}{William~Yang Wang}.} \bibinfo{year}{2018}\natexlab{b}.
\newblock \showarticletitle{Video Captioning via Hierarchical Reinforcement
  Learning}. In \bibinfo{booktitle}{\emph{2018 {IEEE} Conference on Computer
  Vision and Pattern Recognition, {CVPR} 2018, Salt Lake City, UT, USA, June
  18-22, 2018}}. \bibinfo{pages}{4213--4222}.
\newblock


\bibitem[\protect\citeauthoryear{Wang, Luo, Li, Huang, and Yin}{Wang
  et~al\mbox{.}}{2018c}]%
        {DAM}
\bibfield{author}{\bibinfo{person}{Ziwei Wang}, \bibinfo{person}{Yadan Luo},
  \bibinfo{person}{Yang Li}, \bibinfo{person}{Zi Huang}, {and}
  \bibinfo{person}{Hongzhi Yin}.} \bibinfo{year}{2018}\natexlab{c}.
\newblock \showarticletitle{Look Deeper See Richer: Depth-aware Image Paragraph
  Captioning}. In \bibinfo{booktitle}{\emph{2018 {ACM} Multimedia Conference on
  Multimedia Conference, {MM} 2018, Seoul, Republic of Korea, October 22-26,
  2018}}. \bibinfo{pages}{672--680}.
\newblock


\bibitem[\protect\citeauthoryear{Williams}{Williams}{1992}]%
        {reinforce}
\bibfield{author}{\bibinfo{person}{Ronald~J. Williams}.}
  \bibinfo{year}{1992}\natexlab{}.
\newblock \showarticletitle{Simple Statistical Gradient-Following Algorithms
  for Connectionist Reinforcement Learning}.
\newblock \bibinfo{journal}{\emph{Machine Learning}}  \bibinfo{volume}{8}
  (\bibinfo{year}{1992}), \bibinfo{pages}{229--256}.
\newblock


\bibitem[\protect\citeauthoryear{Xu, Ba, Kiros, Cho, Courville, Salakhutdinov,
  Zemel, and Bengio}{Xu et~al\mbox{.}}{2015}]%
        {Att}
\bibfield{author}{\bibinfo{person}{Kelvin Xu}, \bibinfo{person}{Jimmy Ba},
  \bibinfo{person}{Ryan Kiros}, \bibinfo{person}{Kyunghyun Cho},
  \bibinfo{person}{Aaron~C. Courville}, \bibinfo{person}{Ruslan Salakhutdinov},
  \bibinfo{person}{Richard~S. Zemel}, {and} \bibinfo{person}{Yoshua Bengio}.}
  \bibinfo{year}{2015}\natexlab{}.
\newblock \showarticletitle{Show, Attend and Tell: Neural Image Caption
  Generation with Visual Attention}. In \bibinfo{booktitle}{\emph{Proceedings
  of the 32nd International Conference on Machine Learning, {ICML} 2015, Lille,
  France, 6-11 July 2015}}. \bibinfo{pages}{2048--2057}.
\newblock


\bibitem[\protect\citeauthoryear{Xu, Jiang, Qin, Wang, and Du}{Xu
  et~al\mbox{.}}{2018}]%
        {poetry2}
\bibfield{author}{\bibinfo{person}{Linli Xu}, \bibinfo{person}{Liang Jiang},
  \bibinfo{person}{Chuan Qin}, \bibinfo{person}{Zhe Wang}, {and}
  \bibinfo{person}{Dongfang Du}.} \bibinfo{year}{2018}\natexlab{}.
\newblock \showarticletitle{How Images Inspire Poems: Generating Classical
  Chinese Poetry from Images with Memory Networks}. In
  \bibinfo{booktitle}{\emph{Proceedings of the Thirty-Second {AAAI} Conference
  on Artificial Intelligence, (AAAI-18), the 30th innovative Applications of
  Artificial Intelligence (IAAI-18), and the 8th {AAAI} Symposium on
  Educational Advances in Artificial Intelligence (EAAI-18), New Orleans,
  Louisiana, USA, February 2-7, 2018}}. \bibinfo{pages}{5618--5625}.
\newblock


\bibitem[\protect\citeauthoryear{Yang, Lin, Suo, and Li}{Yang
  et~al\mbox{.}}{2018a}]%
        {poetry3}
\bibfield{author}{\bibinfo{person}{Xiaopeng Yang}, \bibinfo{person}{Xiaowen
  Lin}, \bibinfo{person}{Shunda Suo}, {and} \bibinfo{person}{Ming Li}.}
  \bibinfo{year}{2018}\natexlab{a}.
\newblock \showarticletitle{Generating Thematic Chinese Poetry using
  Conditional Variational Autoencoders with Hybrid Decoders}. In
  \bibinfo{booktitle}{\emph{Proceedings of the Twenty-Seventh International
  Joint Conference on Artificial Intelligence, {IJCAI} 2018, July 13-19, 2018,
  Stockholm, Sweden.}} \bibinfo{pages}{4539--4545}.
\newblock
\urldef\tempurl%
\url{https://doi.org/10.24963/ijcai.2018/631}
\showDOI{\tempurl}


\bibitem[\protect\citeauthoryear{Yang, Zhou, Ai, Bin, Hanjalic, Shen, and
  Ji}{Yang et~al\mbox{.}}{2018b}]%
        {yang2}
\bibfield{author}{\bibinfo{person}{Yang Yang}, \bibinfo{person}{Jie Zhou},
  \bibinfo{person}{Jiangbo Ai}, \bibinfo{person}{Yi Bin}, \bibinfo{person}{Alan
  Hanjalic}, \bibinfo{person}{Heng~Tao Shen}, {and} \bibinfo{person}{Yanli
  Ji}.} \bibinfo{year}{2018}\natexlab{b}.
\newblock \showarticletitle{Video Captioning by Adversarial {LSTM}}.
\newblock \bibinfo{journal}{\emph{IEEE Transactions on Image Processing}}
  \bibinfo{volume}{27}, \bibinfo{number}{11} (\bibinfo{year}{2018}),
  \bibinfo{pages}{5600--5611}.
\newblock


\bibitem[\protect\citeauthoryear{Yu, Wang, Huang, Yang, and Xu}{Yu
  et~al\mbox{.}}{2016}]%
        {video}
\bibfield{author}{\bibinfo{person}{Haonan Yu}, \bibinfo{person}{Jiang Wang},
  \bibinfo{person}{Zhiheng Huang}, \bibinfo{person}{Yi Yang}, {and}
  \bibinfo{person}{Wei Xu}.} \bibinfo{year}{2016}\natexlab{}.
\newblock \showarticletitle{Video Paragraph Captioning Using Hierarchical
  Recurrent Neural Networks}. In \bibinfo{booktitle}{\emph{2016 {IEEE}
  Conference on Computer Vision and Pattern Recognition, {CVPR} 2016, Las
  Vegas, NV, USA, June 27-30, 2016}}. \bibinfo{pages}{4584--4593}.
\newblock


\bibitem[\protect\citeauthoryear{Yu}{Yu}{2018}]%
        {efficiency}
\bibfield{author}{\bibinfo{person}{Yang Yu}.} \bibinfo{year}{2018}\natexlab{}.
\newblock \showarticletitle{Towards Sample Efficient Reinforcement Learning}.
  In \bibinfo{booktitle}{\emph{Proceedings of the Twenty-Seventh International
  Joint Conference on Artificial Intelligence, {IJCAI} 2018, July 13-19, 2018,
  Stockholm, Sweden.}} \bibinfo{pages}{5739--5743}.
\newblock


\bibitem[\protect\citeauthoryear{Zhang, Sung, Liu, Xiang, Gong, Yang, and
  Hospedales}{Zhang et~al\mbox{.}}{2017}]%
        {ac}
\bibfield{author}{\bibinfo{person}{Li Zhang}, \bibinfo{person}{Flood Sung},
  \bibinfo{person}{Feng Liu}, \bibinfo{person}{Tao Xiang},
  \bibinfo{person}{Shaogang Gong}, \bibinfo{person}{Yongxin Yang}, {and}
  \bibinfo{person}{Timothy~M Hospedales}.} \bibinfo{year}{2017}\natexlab{}.
\newblock \showarticletitle{Actor-Critic Sequence Training for Image
  Captioning}. In \bibinfo{booktitle}{\emph{NIPS Workshop on Visually-Grounded
  Interaction and Language}}.
\newblock


\bibitem[\protect\citeauthoryear{Zhang, Yang, Zhang, Ji, Shen, and Chua}{Zhang
  et~al\mbox{.}}{2019}]%
        {yang}
\bibfield{author}{\bibinfo{person}{Mingxing Zhang}, \bibinfo{person}{Yang
  Yang}, \bibinfo{person}{Hanwang Zhang}, \bibinfo{person}{Yanli Ji},
  \bibinfo{person}{Heng~Tao Shen}, {and} \bibinfo{person}{Tat-Seng Chua}.}
  \bibinfo{year}{2019}\natexlab{}.
\newblock \showarticletitle{More is Better: Precise and Detailed Image
  Captioning using Online Positive Recall and Missing Concepts Mining}.
\newblock \bibinfo{journal}{\emph{IEEE Transactions on Image Processing}}
  \bibinfo{volume}{28}, \bibinfo{number}{1} (\bibinfo{year}{2019}),
  \bibinfo{pages}{32--44}.
\newblock


\end{thebibliography}
}
\end{document}